\documentclass{article}

\usepackage[T1]{fontenc}
\usepackage{iclr2027_conference,times}
\iclrfinalcopy 

\usepackage[utf8]{inputenc}
\usepackage[hidelinks]{hyperref}
\usepackage{url}
\usepackage{booktabs}
\usepackage{multirow}
\usepackage{amsfonts}
\usepackage{nicefrac}
\usepackage{microtype}
\usepackage{xcolor}
\usepackage{graphicx}
\usepackage{float}
\usepackage{tabularx}
\usepackage{listings}
\usepackage[most]{tcolorbox}
\tcbuselibrary{listings,breakable}
\usepackage{amsmath}
\usepackage{amssymb}
\usepackage{mathtools}
\usepackage{amsthm}
\usepackage{fvextra}
\usepackage[capitalize,noabbrev]{cleveref}

\makeatletter
\@ifundefined{crefname}{}{%
  \crefname{blocks}{Box}{Boxes}%
  \Crefname{blocks}{Box}{Boxes}%
  \crefname{interfaces}{Interface}{Interfaces}%
  \Crefname{interfaces}{Interface}{Interfaces}%
}
\makeatother

\lstdefinestyle{leanappendix}{
  basicstyle=\scriptsize\ttfamily,
  breaklines=true,
  breakatwhitespace=false,
  columns=fullflexible,
  keepspaces=true,
  showstringspaces=false,
  literate=
    {ℝ}{{$\mathbb{R}$}}1
    {ℂ}{{$\mathbb{C}$}}1
    {ℕ}{{$\mathbb{N}$}}1
    {→}{{$\to$}}1
    {↔}{{$\leftrightarrow$}}1
    {∀}{{$\forall$}}1
    {∃}{{$\exists$}}1
    {∈}{{$\in$}}1
    {∧}{{$\land$}}1
    {∨}{{$\lor$}}1
    {≠}{{$\ne$}}1
    {≤}{{$\le$}}1
    {≥}{{$\ge$}}1
    {⊆}{{$\subseteq$}}1
    {∪}{{$\cup$}}1
    {⊥}{{$\bot$}}1
    {≃}{{$\simeq$}}1
    {×}{{$\times$}}1
    {•}{{$\bullet$}}1
    {≫}{{$\gg$}}1
    {↥}{{$\uparrow$}}1
    {‖}{{$\|$}}1
    {⦃}{{$\{\!\{$}}1
    {⦄}{{$\}\!\}$}}1
    {«}{{\guillemotleft}}1
    {»}{{\guillemotright}}1
    {·}{{$\cdot$}}1
    {γ}{{$\gamma$}}1
    {₀}{{$_0$}}1
    {₁}{{$_1$}}1
    {₂}{{$_2$}}1
    {∑}{{$\sum$}}1
    {σ}{{$\sigma$}}1
    {⁻¹}{{$^{-1}$}}2
}

\newcommand{\mainResultsFigWidth}{0.96\linewidth}
\newcommand{\mainResultsFigHeight}{0.40\textheight}

\newif\ifshowchanges\showchangesfalse

\title{Beyond Compilation: Evaluating Faithful\\ Natural-Language-to-Lean Statement\\ Formalization}

\author{Ke Zhang$^{1}$, Patricio Gallardo Candela$^{1*}$, Sudhir Murthy$^{1}$, \\ \bf Yi Xie$^{2}$, Zhi Wang$^{3}$, Maziar Raissi$^{1}$\thanks{Corresponding authors: Maziar Raissi and Patricio Gallardo Candela.} \\
\normalfont\small $^{1}$University of California, Riverside \quad
$^{2}$University of Arizona \quad
$^{3}$University of California, San Diego \\
\normalfont\small\texttt{\{kzhang153,patricio.gallardocandela,smurt002,maziar.raissi1\}@ucr.edu} \\
\normalfont\small\texttt{yix@arizona.edu}, \texttt{zhw119@ucsd.edu}}

\begin{document}

\maketitle
\lhead{} 

\begin{abstract}
Lean verifies that a generated declaration is well typed, but cannot check whether it states what the user intended. We measure how far compilation overstates faithfulness in statement autoformalization without canonical Lean targets. Our criterion requires Lean compilation and agreement of two LLM judges from different model families. On an independent random audit, it agrees with the human majority on 91.5\% of cases; it rarely accepts an unfaithful output but rejects some faithful ones. Across eight systems on 227 university-level statements, the gap between compiling and accepted outputs ranges from 1.3 to 29.5 percentage points: our tool-augmented agent compiles 87.2\% of the statements, but only 57.7\% are accepted. Human review of all the agent's rejections confirms that 23.8 of its 29.5 points are real translation errors. For the gaps of the seven one-shot systems, a two-stage review by Claude Opus 5.5 with Lean and GPT-6.1 Sol, validated on these human labels, finds that 78--100\% are real translation errors for six of them. When GPT-5.2, one of the two judges, grades the translations of two models released after it, GPT-6 Astra and Claude Opus 5.5, it accepts 83\% of the statements for each, against 21\% for one-shot GPT-5.2. For these two models, the same review classifies only 4 of their 54 rejections (7\%) as translation errors, so most of their measured gap is judge errors rather than translation errors. This is consistent with these systems erring less while the judge does not: the fewer errors a system makes, the smaller the share of its rejections that are translation errors. Finally, a $2^3$ tool ablation of the agent shows that Lean feedback drives most of its compilation gain, and that nearly half of this gain fails the semantic criterion. Code and data are available at \url{https://github.com/grenadeComing/beyond-compilation}.
\end{abstract}

\section{Introduction} \label{sec:intro}

\begin{figure}[t]
    \centering
    \includegraphics[width=\mainResultsFigWidth,height=\mainResultsFigHeight,keepaspectratio]{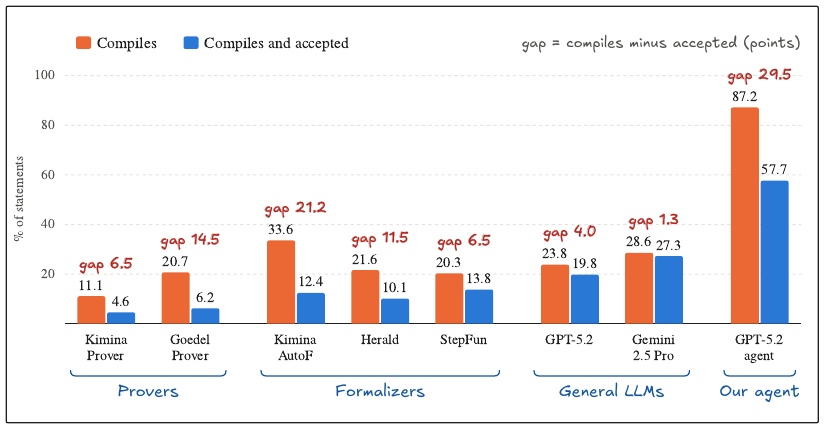}
    \caption{\textbf{Compilation and acceptance by the criterion across systems.} Every system has a gap; it ranges from 1.3 to 29.5 percentage points on the 227 statements. Table~\ref{tab:gap_real} gives how much of each gap is translation errors. Kimina Prover, Kimina AutoF, and StepFun are evaluated on the 217 statements whose recorded input matches the benchmark statement.}
    \label{fig:main_results}
\end{figure}

Proof assistants such as Lean~4~\citep{deMouraUllrich2021lean4} provide a trusted kernel for checking formal syntax, types, and proofs. In theorem proving, the target statement is fixed, so checker acceptance is a strong success signal~\citep{zheng2022minif2f,yang2023leandojo}. In statement autoformalization, however, the model generates the target itself: the generated Lean~4 code may compile while omitting a hypothesis, changing a domain, or strengthening or weakening the conclusion. Lean certifies the theorem the model wrote, not that it wrote the theorem the user meant.

Existing work evaluates statement translation with reference pairs, back-translation, semantic scoring, human annotation, alignment models, LLM-judge ensembles, or formal equivalence checks~\citep{azerbayev2023proofnet,ying2024lean,gaoherald,lu2025formalalign,yu2025mathesis,zhang2025beyondgold,liu2025rethinking}. The difficulty that remains is evaluating open-ended mathematical statements when several Lean encodings may be faithful and no canonical declaration exists: compilation then measures only formal validity, and exact-match or reference-based evaluation is unavailable. Faithfulness is then scored by human reviewers or a reference-free judge, so the measured gap between compiling and accepted outputs mixes the system's real translation errors with two other kinds of rejection: misjudgments by the judge, and rejections on statements that are imprecise or ambiguous as written.

We study a reference-free criterion: an output is accepted if it compiles and two LLM judges from different families, GPT-5.2 and Gemini-2.5-Pro, both grade it at least 9 of 10 for faithfulness (Section~\ref{subsec:evaluation_protocol}). We ask three empirical questions. \textbf{Q1: How closely does this criterion agree with human review, and on which side does it err?} \textbf{Q2: Is the compile--faithfulness gap systematic across model families, and how much does its size vary?} \textbf{Q3: How much of each measured gap is translation errors?} We answer Q1 with blinded human audits, a same-sample comparison to LeanScorer~\citep{yu2025mathesis}, a third-family judge, threshold sensitivity, and a BEq formal cross-check. We answer Q2 by applying one criterion to general LLMs, specialized formalizers, prover-oriented models, and tool-augmented agents on 227 university-level statements. We answer Q3 with human labels on every rejection of our agent and, for the other translation systems, with a two-stage review of each rejected compiling output that we first validate on these human labels. Q1 and Q3 thus follow the same logic: an automatic evaluator is first compared with human labels and then applied to systems that have none.

On an independent random audit, the criterion agrees with the human majority on 54/59 cases (91.5\%, 95\% CI: 81.6--96.3\%), and where two reviewers agree, it matches them about as often as the third reviewer does. Every system shows a compile--faithfulness gap, ranging from 1.3 to 29.5 percentage points (Figure~\ref{fig:main_results}); our tool-augmented agent compiles 87.2\% of the statements, but our criterion accepts only 57.7\%. For seven of the eight systems, most of the gap is real translation errors (Table~\ref{tab:gap_real}). Human review of all the agent's rejections confirms 23.8 of its 29.5 points. For the seven one-shot translation systems, a two-stage review by Claude Opus 5.5 with access to Lean and GPT-6.1 Sol, which matches agreeing human reviewers at least as often as the third reviewer does on 69 human-labeled rejections, finds that 78--100\% of the gap is translation errors for six of them. When GPT-5.2 grades the translations of two models released after it, GPT-6 Astra and Claude Opus 5.5, the share is small: the review classifies only 4 of their 54 rejections as translation errors, and Lean proves nine pairs of differently graded outputs equivalent (Section~\ref{sec:newer}).

Finally, a $2^3$ tool ablation of the agent (Section~\ref{sec:factorial}) shows that Lean feedback drives most of its compilation gain, and that nearly half of this gain fails the semantic criterion. We release the benchmark with its 18 defective statements flagged, every system's outputs and judge grades, the human labels, the reviews of all 268 reviewed rejections with their Lean checks, the evaluation scripts, and the agents' tool-call logs.

\section{Evaluation without Canonical Lean Targets}
\label{sec:problem}

\subsection{Task Definition}
Let $\mathcal{X}$ be the space of informal mathematical statements and $\mathcal{Y}$ the space of Lean~4 declarations. Given an input $x \in \mathcal{X}$, a system generates a statement $y \in \mathcal{Y}$ that should satisfy three criteria: \textbf{syntactic validity} ($y$ compiles in Lean~4 with Mathlib), \textbf{statement-centricity} ($y$ omits proofs via \texttt{:= by sorry}), and \textbf{semantic faithfulness} ($y$ expresses the same mathematical claim as $x$).

\subsection{Benchmark}
We collect university-level statements from four open-source LaTeX textbooks, three written for undergraduate courses and one, the complex analysis text, for a first graduate course: Lebl's \textit{Basic Analysis}~\citep{lebl_basic_analysis_github} for Real Analysis (48 statements), Lebl's \textit{Guide to Cultivating Complex Analysis}~\citep{lebl_ca_analysis_github} for Complex Analysis (19), McKay's \textit{Notes on Topology}~\citep{mckay_topology_notes_github} for Topology (63), and Doty's \textit{Abstract Algebra}~\citep{doty_abstract_algebra_notes_github} for Algebra (97), 227 statements in total. Each statement is taken from the source text it cites, and imprecisions in the sources are kept as written. The sources contain no formal code, so there is no canonical Lean target, and a statement may have several faithful Lean encodings. The release flags 18 statements that are false as written, as in their sources, or lost context during extraction, and Appendix~\ref{app:flagged} repeats the main measurements without them.

\subsection{Evaluated Systems}
We evaluate one-shot general LLMs (GPT-5.2, Gemini-2.5-Pro), specialized statement formalizers (Herald~\citep{gaoherald}, Kimina-Autoformalizer~\citep{kimina_autoformalizer_card}, StepFun-Formalizer~\citep{wu2025stepfunformalizer}), prover-oriented Lean models (Kimina-Prover~\citep{wang2025kimina}, Goedel-Prover~\citep{lin2025goedelproverv2}), and a tool-augmented GPT-5.2 agent with drafting, search, and Lean REPL feedback tools, all in the same statement-only setting. Section~\ref{sec:factorial} describes the agent's tools and varies them in a factorial design. We also run Aristotle~\citep{harmonic2025aristotle}, Harmonic's Lean-based formal reasoning system, on 60 randomly sampled benchmark statements. We score its outputs with our criterion, and three human reviewers score the same outputs independently; this forms the external audit in Section~\ref{sec:robustness} (Batch~B). Because Aristotle covers only this sample, it is not included in Figure~\ref{fig:main_results}. Section~\ref{sec:newer} adds two models released after GPT-5.2, GPT-6 Astra and Claude Opus 5.5, run one-shot with the same prompt.

\subsection{Evaluation Protocol}
\label{subsec:evaluation_protocol}
We compile each output in Lean~4 with Mathlib. For compiling outputs, GPT-5.2 and Gemini-2.5-Pro independently compare the natural-language source with the generated declaration and assign integer faithfulness grades from 0 to 10 under a shared rubric (Appendix~\ref{app:judge_rubric}); grades 9--10 indicate matching meaning up to a negligible discrepancy. An output is \textbf{accepted} by the criterion if it compiles and both judges grade it at least 9:
\[
\mathrm{Accepted}(x,y)=\mathrm{Compiles}(y)\land
\mathrm{grade}_{\mathrm{GPT}}(x,y)\geq9\land
\mathrm{grade}_{\mathrm{Gemini}}(x,y)\geq9.
\]
We validate this strict operating point against human judgments in Section~\ref{sec:robustness}. A \textbf{rejection} is a compiling output that the criterion does not accept, and a system's \textbf{measured gap} is its compile rate minus its acceptance rate, that is, its rejections in percentage points of the statements. A rejection is either a translation error of the system or is caused by a misjudgment of the judge or by an imprecise or ambiguous statement; Section~\ref{sec:decomp} separates them.

\paragraph{Cross-judge consensus.}
Gemini-2.5-Pro accepts at least as many outputs as GPT-5.2 for every system, and almost every GPT-5.2 positive is also a Gemini positive. Requiring both judges therefore removes few GPT-accepted outputs: across all systems, 1309 of 1326 GPT-5.2 positives (98.7\%) are also Gemini positives (per-system counts in Appendix Table~\ref{tab:judge_comparison_full}; domain-level disagreement in Appendix Figure~\ref{fig:judge_disagreement_domain}). Their agreement is not independent confirmation, since the errors of different LLMs are correlated~\citep{goel2025great,kim2025correlated}.

\section{\texorpdfstring{Measuring the Compile--Faithfulness Gap}{Measuring the Compile--Faithfulness Gap}}
\label{sec:benchmarking}

\subsection{\texorpdfstring{Q1: How Closely Does the Criterion Agree with Human Review, and on Which Side Does It Err?}{Q1: How Closely Does the Criterion Agree with Human Review, and on Which Side Does It Err?}}
\label{sec:robustness}

We compare the criterion with human review in three audits (Table~\ref{tab:human_calibration_main}). The positive audit and Batch A cover the two sides of the criterion's decision boundary on our agent's outputs: an expert reviews a domain-proportional random sample of outputs that GPT-5.2 accepts (70 of the 72 scored are also accepted by Gemini), and three blinded reviewers review every compiling output the criterion rejects. Batch B is a full evaluation of an external system: the same three blinded reviewers score Aristotle's outputs for the 60 sampled statements (26 Algebra, 12 Real Analysis, 4 Complex Analysis, 18 Topology), covering outputs the criterion accepts and outputs it rejects; one missing output leaves 59 cases, all with a human-majority label.

\begin{table}[t]
\centering
\caption{\textbf{Human audits of the compile + GPT$\cap$Gemini criterion.} Intervals are Wilson 95\% CIs.}
\label{tab:human_calibration_main}
\small
\setlength{\tabcolsep}{4pt}
\renewcommand{\arraystretch}{1.15}
\begin{tabularx}{\linewidth}{@{}>{\raggedright\arraybackslash\leavevmode}p{0.13\linewidth}>{\raggedright\arraybackslash\leavevmode}p{0.26\linewidth}>{\raggedright\arraybackslash\leavevmode}p{0.24\linewidth}>{\raggedright\arraybackslash}X@{}}
\toprule
\textbf{Audit} & \textbf{System and selection} & \textbf{Human coverage} & \textbf{Result (95\% CI)} \\
\midrule
Positive region & Our agent; random sample of outputs that GPT-5.2 accepts. & One expert; 72 scored, 70 of them also accepted by Gemini. & Precision: 67/70 = 95.7\% \textbf{[88.1, 98.5]\%}. \\
\midrule
Batch A: negative region & Same agent; all 67 compile-pass outputs rejected by strict consensus. & Three blinded reviewers; 67/67 have a majority label. & Negative confirmation: 54/67 = 80.6\% \textbf{[69.6, 88.3]\%}; 13 are judged faithful. \\
\midrule
Batch B: full evaluation & Aristotle; 60 randomly sampled benchmark statements; realized domains 26/12/4/18. & Three blinded reviewers; 1 missing output leaves 59 effective; 59 have a majority label. & Agreement: 54/59 = 91.5\% \textbf{[81.6, 96.3]\%}. Precision: 44/46 = 95.7\% \textbf{[85.5, 98.8]\%}. Negative confirmation: 10/13 = 76.9\% \textbf{[49.7, 91.8]\%}. \\
\bottomrule
\end{tabularx}
\end{table}

On the random sample, the criterion agrees with the human majority on 91.5\% of cases (95\% CI: 81.6--96.3\%); on the 53 compiling outputs, where only the judges decide, it agrees on 48 (90.6\%; 95\% CI: 79.7--95.9\%). The reviewers also disagree with one another, agreeing pairwise on 75.9--91.1\% of the cases; where two of them agree, the criterion matches them on 92.7--96.0\%, and the third reviewer on 80.0--97.6\%. Agreement is asymmetric: humans confirm 95.7\% of automatic positives in both audits but 76.9--80.6\% of automatic negatives, and 4 of Aristotle's 7 compiling negatives, so the criterion rarely accepts an unfaithful output but rejects some faithful ones. For example, every reviewer who scored it accepts an output that states that every contractible space is connected and has trivial homotopy groups, through Mathlib's \texttt{ContractibleSpace} and \texttt{HomotopyGroup}, but Gemini grades it 5. Conversely, both judges give 10 to an output that states the injectivity of Mathlib's bundled homomorphism \texttt{MonoidHom.inl}, $x\mapsto(x,1)$, whose homomorphism property comes with its type; two of three reviewers reject it and one accepts it, a disagreement among humans over whether a property supplied by the type counts as stated (counted as faithful, it would raise agreement to 55/59 and precision to 45/46). Appendix~\ref{app:negative_audit} reports the review protocol and case studies.

\paragraph{Same-sample comparison with LeanScorer.}
On the same Batch-B sample, our rule agrees with the human majority on 54/59 cases, versus 45/59 for the public two-stage LeanScorer rule (DeepSeek-V3-0324, threshold 0.6)~\citep{yu2025mathesis} (Table~\ref{tab:metric_validation}). The paired outcomes are 43 both correct, 11 ours only, 2 LeanScorer only, and 3 both wrong (exact McNemar $p=0.0225$).

\begin{table}[t]
\centering
\caption{\textbf{Evaluator comparison on Batch B.} Human-majority faithfulness is the comparison label; both rules use the same compilation results.}
\label{tab:metric_validation}
\footnotesize
\setlength{\tabcolsep}{5pt}
\begin{tabularx}{\linewidth}{@{}l>{\centering\arraybackslash}X>{\centering\arraybackslash}X>{\centering\arraybackslash}X@{}}
\toprule
\textbf{Rule} & \textbf{Human agreement} & \textbf{Positive precision} & \textbf{Negative confirmation} \\
\midrule
Compile + GPT$\cap$Gemini & 54/59 (91.5\%) & 44/46 (95.7\%) & 10/13 (76.9\%) \\
Compile + LeanScorer & 45/59 (76.3\%) & 36/39 (92.3\%) & 9/20 (45.0\%) \\
\bottomrule
\end{tabularx}
\end{table}

\paragraph{Independent judge and threshold robustness.}
We apply the same rubric and threshold to Claude Sonnet~5, excluding unresolved API calls from the denominators (Table~\ref{tab:third_judge_main}). Sonnet agrees with the primary rule on 89.2\% of our agent's resolved outputs and 87.4\% across all eight tool configurations of the agent (Section~\ref{sec:factorial}). Varying the GPT/Gemini threshold from 7 to 10 preserves the ordering of factorial main effects, Feedback $>$ Search $>$ Translation, with effects of +21.7--30.9, +6.9--8.5, and +1.0--2.1 points, and keeps our agent's gap between 24.7 and 37.9 points.

\begin{table}[t]
\centering
\caption{\textbf{Third-judge robustness.} The final two columns split agreement by the primary label.}
\label{tab:third_judge_main}
\footnotesize
\setlength{\tabcolsep}{4pt}
\begin{tabularx}{\linewidth}{@{}l>{\centering\arraybackslash}X>{\centering\arraybackslash}X>{\centering\arraybackslash}X>{\centering\arraybackslash}X@{}}
\toprule
\textbf{Scope} & \textbf{Coverage} & \textbf{Agreement} & \textbf{Positive confirmed} & \textbf{Negative confirmed} \\
\midrule
All 8 tool configurations & 1123/1146 (98.0\%) & 981/1123 (87.4\%) & 712/774 (92.0\%) & 269/349 (77.1\%) \\
Our agent & 195/198 (98.5\%) & 174/195 (89.2\%) & 120/131 (91.6\%) & 54/64 (84.4\%) \\
\bottomrule
\end{tabularx}
\end{table}

\paragraph{Formal cross-check with BEq.}
Without a canonical Lean target, we use outputs confirmed by a human as references. As anchors, we take 40 accepted outputs of our agent that the expert confirmed in the positive audit (19 Algebra, 13 Real Analysis, 5 Topology, 3 Complex Analysis). For each anchor, BEq checks whether the accepted outputs for the same statement from the agent's seven other tool configurations (Section~\ref{sec:factorial}) are equivalent to it, 183 pairs in all. Adapted BEq-Normal~\citep{liu2025rethinking} requires Lean-verified implications in both directions and checks that each proof uses its source statement. We run \texttt{exact?} first (105 certificates), then InternLM2-Math-Plus-20B~\citep{ying2024internlmmath} with the published five-shot Normal-filter procedure (eight attempts, 512-token cap), which adds 8. BEq certifies 113 of the 183 pairs as equivalent (61.7\%; anchor-cluster bootstrap 95\% CI: 48.9--73.3\%), including 67 pairs with different normalized encodings (Table~\ref{tab:beq_main}); the other 70 are unresolved by bounded proof search.

\begin{table}[t]
\centering
\caption{\textbf{BEq cross-check on accepted outputs.} Pairs of accepted outputs for the same statement that BEq certifies as equivalent.}
\label{tab:beq_main}
\footnotesize
\setlength{\tabcolsep}{6pt}
\begin{tabularx}{\linewidth}{@{}l>{\centering\arraybackslash\leavevmode}X>{\centering\arraybackslash\leavevmode}X@{}}
\toprule
\textbf{Comparison subset} & \textbf{Certified equivalent} & \textbf{Not established} \\
\midrule
All accepted pairs & \textbf{113/183 (61.7\%)} & 70/183 (38.3\%) \\
Same normalized encoding & \textbf{46/46 (100\%)} & 0/46 (0\%) \\
Different normalized encodings & \textbf{67/137 (48.9\%)} & 70/137 (51.1\%) \\
\bottomrule
\end{tabularx}
\end{table}

Human review measures agreement with the intended meaning, Sonnet tests dependence on the judge family, and BEq supplies formal certificates on a subset. Together they support the reliability of accepted outputs and locate where the criterion is conservative. Rejections of faithful outputs also count toward the measured gap; Section~\ref{sec:decomp} measures how much of each gap is translation errors.

\subsection{Q2: How Large Is the Compile--Faithfulness Gap?}
\label{sec:q2}

Figure~\ref{fig:main_results} compares seven one-shot systems and our agent on the 227 statements. Every system compiles more often than the criterion accepts, but the gap varies: 1.3--21.2 percentage points for the one-shot systems and 29.5 points for our agent. With Sonnet 4.5 and Gemini-2.5-Pro as orchestrators, the agent's gap is 26.9 and 11.0 points (Appendix~\ref{app:multi_model}). Acceptance rates are \textbf{19.8--27.3\%} for general-purpose one-shot LLMs, \textbf{10.1--13.8\%} for specialized formalizers, and \textbf{4.6--6.2\%} for prover-oriented models, compared with \textbf{57.7\%} for our agent.

Despite an explicit statement-only prompt requiring declarations that end in \texttt{:= by sorry}, prover-oriented models frequently emit proof-oriented reasoning, tactic bodies, malformed declarations, or no extractable statement. In these two models, proof-oriented Lean competence does not transfer to prompt-following, faithful statement formalization when the target statement is not supplied.

Our agent shows the practical risk most clearly: 198/227 outputs compile (87.2\%), but only 131/227 (57.7\%) satisfy the semantic criterion. The human majority confirms 54/67 audited compile-pass rejections as genuine mismatches (80.6\%, 95\% CI: 69.6--88.3\%), so at least 54/227 = 23.8 of the 29.5 points are human-confirmed errors of the system, and humans judge the other 13 rejected outputs (5.7 points) faithful. Table~\ref{tab:compile_pass_examples} shows two of the confirmed errors. Compiler-guided repair can produce valid Lean that no longer states the intended theorem.

\begin{table}[t]
\centering
\caption{\textbf{Two compiling outputs of our agent that the human majority rejects} (Batch A). Statements and code are abridged; the full items are in Appendix~\ref{app:human_check_cases}.}
\label{tab:compile_pass_examples}
\footnotesize
\renewcommand{\arraystretch}{1.15}
\begin{tabularx}{\linewidth}{@{}>{\raggedright\arraybackslash}p{0.25\linewidth}>{\raggedright\arraybackslash}p{0.33\linewidth}>{\raggedright\arraybackslash}X@{}}
\toprule
\textbf{Statement (abridged)} & \textbf{Lean output (abridged)} & \textbf{What changed} \\
\midrule
\textbf{A-005.} An analytic $g\colon Z\to\mathbb{C}$ on a domain $Z$ has a logarithm iff (1) $g\neq0$ on $Z$ and (2) $g$ takes every loop in $Z$ to a null-homotopic loop in $\mathbb{C}\setminus\{0\}$. &
Condition (2) becomes $\forall x\in Z$, $\forall\gamma : \texttt{Path}\ x\ x$, $\exists\gamma_0 : \texttt{Path}\ (g\,x)\ (g\,x)$, \texttt{Path.Homotopic} $\gamma_0$ (\texttt{Path.refl}~\_). &
The condition never uses $\gamma$, and $\gamma_0=\texttt{Path.refl}$ satisfies it, so condition (2) always holds. GPT/Gemini: 4/4. Reviewers: 5, 4, 3. \\
\midrule
\textbf{A-030.} A nonempty set $G$ of real $n\times n$ matrices that is closed under products and inverses contains the identity matrix. &
Hypotheses $A\in G\to B\in G\to AB\in G$ and $A\in G\to A^{-1}\in G$; conclusion $1\in G$. &
Mathlib defines $A^{-1}=0$ for singular $A$, so $G=\{0\}$ satisfies every hypothesis but does not contain $1$ (for $n\geq1$): the statement drops the invertibility that a matrix group assumes. GPT/Gemini: 6/10. Reviewers: 10, 5, 2. \\
\bottomrule
\end{tabularx}
\end{table}

With GPT-5.2, Sonnet 4.5, or Gemini-2.5-Pro as its orchestrator, our agent reaches 58--65\% acceptance, against one-shot baselines of 20--33\% for the same models (Appendix~\ref{app:multi_model}).

\section{Q3: How Much of Each Gap Is Translation Errors?}
\label{sec:decomp}

A rejection of our criterion is either a translation error (T), where the output does not state the source statement, or it has another cause. The judge may misjudge a faithful output (J), or the benchmark statement may be defective (B): it is false as written, lost context in extraction, or is ambiguous, so that humans and LLMs can read it differently. For such a statement, an output can only translate it literally or repair it on its own, and the judges may reject either; the rejection is then neither clearly the system's error nor clearly the judge's. Some rejections have both causes (J/B). The measured gap thus consists of translation errors, judge errors, and rejections on defective statements. Of the eight translation systems, only our agent has all its rejected translations audited by humans: 54 of its 67 rejections, 23.8 of its 29.5 points, are human-confirmed translation errors (Section~\ref{sec:q2}). For the other systems, which have no human audits, we decompose the gap with a two-stage review of every rejected compiling output: Claude Opus 5.5, run as agents with access to Lean, classifies each rejection as T, J, or B, and GPT-6.1 Sol then checks each classification. We first run this review on the rejections that humans labeled, those of our agent and of Aristotle, and compare its labels with the human ones (Section~\ref{sec:review}); we then apply it to the seven one-shot systems (Section~\ref{sec:gap_real}) and to two models released after GPT-5.2 (Section~\ref{sec:newer}).

\subsection{A Two-Stage Review Validated on Human Labels}
\label{sec:review}

\paragraph{Procedure.} In the first stage, Claude Opus 5.5 agents, run in Claude Code with access to the Lean REPL and the textbook sources, classify each rejected output as T, J, or B and check in Lean the definitions or conventions that decide a case; they see the statement, the output, and the judges' grades and stated reasons, but no human label. In the second stage, GPT-6.1 Sol reviews each output with the Claude Code review in view. Where the two stages agree, their class is final; where they differ only between J and B, the class is J/B; where exactly one finds a translation error, GPT-6.1 Sol decides in a discussion round that sees both reviews. This decision matches the second stage in 27 of the 28 discussed outputs across all our review runs, so the final class is close to GPT-6.1 Sol's, and we also report translation errors flagged by either stage. Appendix~\ref{app:review_validation} gives the inputs; the prompts, both reviews, the discussions, and the reviewers' Lean checks are released.

\paragraph{Validation.} We ran the review workflow on the 69 compiling outputs that GPT-5.2 rejects and that humans labeled (64 of our agent and 5 of Aristotle). Of the 57 that the human majority judges unfaithful, the two-stage review classifies 47 as translation errors (82.5\%; 95\% CI: 70.6--90.2\%), so its final class misses about one error in six, and 54 counting either stage; of the 12 that the majority judges faithful, it leaves 8 unflagged (66.7\%; 95\% CI: 39.1--86.2\%) (Table~\ref{tab:review_validation}). Two of the four human-faithful outputs that it flags are errors that Lean confirms and that the majority missed: one drops a hypothesis $a\le b$ without which the statement is false, and one asks for an isomorphism with a monoid that no group has (A-008 and A-094; Appendix~\ref{app:review_validation}); of the eight it leaves unflagged, it attributes four rejections to the judge and four to defective statements.

The human reviewers also disagree with each other, so we compare the review with each of them (Table~\ref{tab:review_parity}). On the outputs where two reviewers agree, the review matches them on 84--91\%, and the held-out third reviewer on 77--84\%. Over all 69 outputs, the review agrees with the human majority on 79.7\% ($\kappa=0.41$), while pairs of reviewers agree with each other on 69.6--74.6\% ($\kappa=0.12$--$0.27$). On these rejected outputs, the review thus agrees with the human reviewers at least as well as they agree with one another, which is the basis for applying it to the systems without human audits.

\begin{table}[t]
\centering
\caption{\textbf{The review against the human majority} on the 69 validation outputs. Errors found: human-unfaithful outputs that the review classifies as translation errors. Faithful left unflagged: human-faithful outputs that it does not. GPT-6.1 Sol alone: a separate run of GPT-6.1 Sol without the first review. Wilson 95\% CIs.}
\label{tab:review_validation}
\small
\setlength{\tabcolsep}{8pt}
\begin{tabular}{lcc}
\toprule
\textbf{Rule} & \textbf{Errors found} & \textbf{Faithful left unflagged} \\
\midrule
Final class T & 47/57 [70.6, 90.2] & 8/12 [39.1, 86.2] \\
T by either stage & 54/57 [85.6, 98.2] & 5/12 [19.3, 68.0] \\
GPT-6.1 Sol alone & 46/57 [68.7, 88.9] & 8/12 [39.1, 86.2] \\
\bottomrule
\end{tabular}

\vspace{10pt}
\caption{\textbf{Agreement with reviewers who agree with each other.} For each held-out reviewer, the outputs on which the other two agree, and how often the held-out reviewer and the review match them (agreement, Cohen's $\kappa$).}
\label{tab:review_parity}
\small
\setlength{\tabcolsep}{8pt}
\begin{tabular}{lrcc}
\toprule
\textbf{Reference} & $n$ & \textbf{Held-out reviewer} & \textbf{Review} \\
\midrule
R3 and R4 agree & 47 & R2: 76.6\%, 0.18 & 89.4\%, 0.39 \\
R2 and R4 agree & 45 & R3: 80.0\%, 0.23 & 91.1\%, 0.56 \\
R2 and R3 agree & 43 & R4: 83.7\%, 0.28 & 83.7\%, 0.44 \\
\bottomrule
\end{tabular}
\end{table}

\subsection{The Gap of Every System}
\label{sec:gap_real}

\begin{table}[t]
\centering
\caption{\textbf{How much of each gap in Figure~\ref{fig:main_results} is translation errors.} Gap: compiling outputs that the criterion rejects, in percentage points of the statements, as in Figure~\ref{fig:main_results}. \textbf{Real gap}: the part of the gap that is translation errors, by the human majority for our agent (Batch~A covers all of its rejections) and by the review of Section~\ref{sec:review} for the one-shot systems. Share: real gap over gap. $^\dagger$Evaluated on 217 statements.}
\label{tab:gap_real}
\setlength{\tabcolsep}{10pt}
\begin{tabular}{lrrr}
\toprule
\textbf{System} & \textbf{Gap} & \textbf{Real gap} & \textbf{Share} \\
\midrule
Kimina-Prover$^\dagger$ & 6.5 & \textbf{6.5} & 100\% \\
Goedel-Prover & 14.5 & \textbf{14.1} & 97\% \\
Kimina-AutoF$^\dagger$ & 21.2 & \textbf{20.3} & 96\% \\
Herald & 11.5 & \textbf{10.1} & 88\% \\
StepFun$^\dagger$ & 6.5 & \textbf{5.1} & 79\% \\
GPT-5.2 & 4.0 & \textbf{3.1} & 78\% \\
Gemini-2.5-Pro & 1.3 & \textbf{0.4} & 33\% \\
\midrule
GPT-5.2 agent & 29.5 & \textbf{23.8} & 81\% \\
\bottomrule
\end{tabular}
\end{table}

The seven one-shot systems have 145 rejections (Figure~\ref{fig:main_results}). The two-stage review classifies 132 of them as translation errors (138 counting either stage; per-system classes in Appendix Table~\ref{tab:earlier_review}). Table~\ref{tab:gap_real} puts these counts on the scale of Figure~\ref{fig:main_results}, in percentage points of the statements. For every system except Gemini-2.5-Pro, one of whose three rejections is a translation error, most of the gap is translation errors: 78--100\%, and 88--100\% for Goedel-Prover, Kimina-Prover, Kimina-Autoformalizer, and Herald. Many of these errors are plain: tautologies, \texttt{True} placeholders, hypotheses that make the statement vacuous, and unrelated algebraic structures on one type. The real gaps order the systems as the measured gaps do. For seven of these eight systems, the gap of Figure~\ref{fig:main_results} is therefore mostly what it appears to be: compiling outputs that do not state the source statement.

\subsection{Two Models Released after the Judge}
\label{sec:newer}

\begin{figure}[t]
\centering
\includegraphics[width=0.62\linewidth]{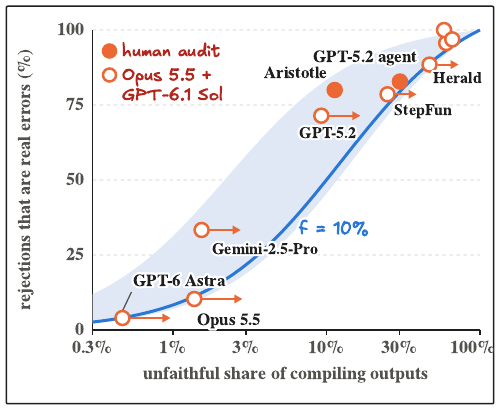}
\caption{\textbf{Share of rejections that are real errors}, against the unfaithful share $e$ of a system's compiling outputs (log scale), with acceptance by GPT-5.2 alone. The band is illustrative: $e(1-m)/(e(1-m)+(1-e)f)$ for $f$ from 2\% to 12\% with $m=10\%$; the solid curve is $f=10\%$. Filled points come from human audits. Open points use the labels of the review of Section~\ref{sec:review}, which covered only rejected outputs: their $e$ counts only the errors found among rejections, so it is a lower bound, and the arrows mark that the true value may lie further right; the later models' overall error rates are not measured. The three open points at the top right are Goedel-Prover, Kimina-Prover, and Kimina-AutoF, drawn without arrows for lack of space.}
\label{fig:judge_regime}
\end{figure}

We ran two models released after GPT-5.2, GPT-6 Astra and Claude Opus 5.5, as two new translation systems, one-shot with the same prompt and Lean environment. For their translations, and in Figure~\ref{fig:judge_regime}, GPT-5.2 is the only judge, since Gemini-2.5-Pro was not available for these runs; on the eight systems, this rule differs from the criterion on at most three statements per system. Without tools, both models compile more statements than our agent (94.3\% and 96.0\%), and GPT-5.2 accepts 83.3\% of the statements for each (20.7\% for one-shot GPT-5.2), leaving gaps of 11.0 and 12.8 points. The source textbooks are public on GitHub, so both models may have seen the statements in training.

The two-stage review classifies 1 of GPT-6 Astra's 25 rejections and 3 of Opus 5.5's 29 as translation errors, or 4 for each counting either stage, so by its labels translation errors make up at most 1.8 of the 11.0 and 12.8 points; since Claude Opus 5.5 runs the first stage, it also reviews its own outputs. The review misses about one error in six (Section~\ref{sec:review}). Since each flagged output is either a real error that the review found or a false flag, its sensitivity bounds the number $X$ of real errors among a system's rejections, whatever its false-flag rate: $X\le T/0.825$ for $T$ flagged outputs, at most 1.2 for GPT-6 Astra and 3.6 for Opus 5.5, if the review finds these models' errors as reliably as those of the validation set. In the clearest error, Opus counts isomorphism classes of group structures on \texttt{Fin n}, but instance resolution compares them through the default multiplication of \texttt{Fin n}, so the statement equates the product of partition numbers with 1, which fails for $n=4$. The review attributes the other 50 rejections to the judge (32), to defective statements (11), or to both (7). By its account, GPT-5.2 rejects equivalent but non-standard encodings, and grades essentially identical outputs of the two models differently, in one case 10 and 2 (Appendix~\ref{app:newer}). Part of this holds without trusting any reviewer: for 9 pairs of outputs that GPT-5.2 grades differently, Lean proves the two statements equivalent.

This contrast with the seven one-shot systems, for which 71--100\% of GPT-5.2's rejections are translation errors for six of the seven, is consistent with a simple relation. If a judge rejects a fraction $f$ of faithful compiling outputs and accepts a fraction $m$ of unfaithful ones, the share of rejections that are real errors, for a system whose compiling outputs are unfaithful at rate $e$, is the positive predictive value of a rejection~\citep{altman1994predictive},
\[
\frac{e(1-m)}{e(1-m)+(1-e)f},
\]
which falls below one half once $e<f/(1-m+f)$, close to $f$ when $m$ is small (Figure~\ref{fig:judge_regime}). With $f=m=10\%$, for example, 79\% of the rejections are real errors for a system whose compiling outputs are 30\% wrong, but 8\% for one whose outputs are 1\% wrong. Our audit does not estimate the later models' overall error rates, since their accepted outputs were not reviewed: the errors found among their rejections are 0.5\% and 1.4\% of their compiling outputs, a lower bound on $e$. Nor does it bear on how systems rank: if all systems shared $f$ and $m$, the accepted share of compiling outputs, $m+(1-f-m)s$ for a faithful share $s$, would still order them correctly. Appendix~\ref{app:newer} reports the judge's failures on these outputs, Lean checks, a second judge, re-grading, supplying definitions to the judge, and the defective statements; Appendix~\ref{app:flagged} repeats the measurements without the 18 flagged statements.

\section{\texorpdfstring{Tool Ablation of the Agent}{Tool Ablation of the Agent}}
\label{sec:factorial}

Aggregate accuracy conflates three questions: whether an output compiles, whether it preserves the intended statement, and how much tool interaction it costs. A tool can raise the compile rate while enlarging the compile-pass but semantically rejected bucket, so we ask which boundary each tool moves: validity, faithfulness, or efficiency. Since most of the agent's gap is translation errors (Table~\ref{tab:gap_real}), this also asks which tools produce the gap. In this section, as in its tables, \emph{faithful} means accepted by the criterion.

\subsection{Agent and Tool Factors}
\label{sec:method}
The agent (Appendix Figure~\ref{fig:agent_logic}) implements three common remedies in autoformalization pipelines in a controlled way: drafting from a Lean-specialized prior, grounding through search, and repair through verifier feedback. A central LLM orchestrator (GPT-5.2; Sonnet 4.5 and Gemini-2.5-Pro in Appendix~\ref{app:multi_model}) interacts with the Lean~4 environment through a defined API and keeps a persistent state of compiler errors, retrieved symbols, and draft translations before choosing its next step. \textbf{Expert Drafting (T)} is a parametric translation prior: the \texttt{lean4\_translation} tool requests a Lean statement draft from the fine-tuned Herald model, which the orchestrator then edits or verifies. \textbf{Knowledge Search (S)} grounds the translation in Mathlib through symbol lookup (\texttt{lean\_inspect\_name}, \texttt{lean\_resolve\_name}), which runs \texttt{\#check} and \texttt{\#print} on identifiers through the Lean REPL, and through general web search; these tools return identifier candidates, types, and definitions. \textbf{Compiler Feedback (F)} exposes Lean elaboration through \texttt{lean4\_repl\_runner}, giving whole-statement validity checks and error messages for syntax, type, namespace, and elaboration repair. Tool definitions and signatures are in Appendix~\ref{app:implementation}.

All configurations share the same role definition and Lean-generation instructions: the agent must translate rather than prove, produce one theorem declaration ending in \texttt{:= by sorry}, import Mathlib, and avoid invented definitions or axioms. Only the tool-availability block changes, declaring which tools the agent may call, so the factorial analysis varies tool access while the task definition, output requirements, and anti-hallucination instructions stay fixed. Full prompt templates are in Appendix~\ref{app:implementation}.

\subsection{Factorial Design over (T,F,S)}
\label{subsec:exp_design}

Each tool setting is a bit vector
\[
c=(t,f,s)\in\{0,1\}^3,\qquad
\mathcal{C}=\{000,001,010,011,100,101,110,111\},
\]
where the bits denote access to (T,F,S). We run every benchmark statement under every $c\in\mathcal{C}$ and evaluate metrics $m(c)$ such as compile rate, consensus-faithful rate, and tool cost. Main effects are high-minus-low averages over the other factors; for example,
\[
\Delta_F(m)=\frac{1}{4}\sum_{t,s\in\{0,1\}}\bigl[m(t,1,s)-m(t,0,s)\bigr].
\]
Configuration 111, with all three tools, is our agent, and 000 is the one-shot GPT-5.2 baseline of Section~\ref{sec:benchmarking}; the other six configurations toggle tool definitions in the system prompt while the task prompt stays fixed.

\subsection{Main Effects: Validity vs. Faithfulness}
\label{subsec:factorial_results}

Table~\ref{tab:factorial_results} reports compile and faithful rates for all configurations, with each row comparing the same $(T,S)$ setting without and with elaboration feedback (\textbf{F}). Main and interaction effects are standard factorial contrasts: differences in mean response between the high and low levels of a factor, averaged over the other factors (Table~\ref{tab:main_effects}).

\begin{table}[t]
\centering
\caption{\textbf{Tool ablation results.} Performance across all $2^3$ configurations ($N=227$). Each row fixes the translation prior (\textbf{T}) and search (\textbf{S}) factors and compares runs without vs.\ with elaboration feedback (\textbf{F}).}
\label{tab:factorial_results}
{\footnotesize
\setlength{\tabcolsep}{5pt}
\renewcommand{\arraystretch}{1.05}
\begin{tabular}{@{}cc|cc|cc@{}}
\toprule
\multicolumn{2}{c|}{\textbf{Fixed factors}} &
\multicolumn{2}{c|}{\textbf{$F=0$}} &
\multicolumn{2}{c}{\textbf{$F=1$}} \\
\textbf{T} & \textbf{S} & \textbf{Comp.} & \textbf{Faith.} & \textbf{Comp.} & \textbf{Faith.} \\
\midrule
0 & 0 & 23.79 & 19.82 & 89.43 & 55.95 \\
1 & 0 & 28.63 & 24.67 & 91.19 & 53.74 \\
0 & 1 & 48.02 & 35.24 & 83.26 & 58.15 \\
1 & 1 & 53.30 & 37.00 & 87.22 & 57.71 \\
\bottomrule
\end{tabular}
}
\end{table}

Tools move different boundaries. Feedback greatly expands validity: $(000)\to(010)$ raises compiled outputs from 54 to 203 and faithful outputs from 45 to 127, but also raises the compile-pass gap from 9 to 76. With feedback enabled, search changes final faithfulness little but reduces the compile-pass gap from 76 to 57 for $T=0$ and from 85 to 67 for $T=1$. The full conditional table and tool-usage counts are in Appendix~\ref{app:factorial_extra}.

\begin{table}[t]
\centering
\caption{\textbf{Main effects of each tool.} Average high-minus-low differences over the other two factors, in percentage points, with 95\% CIs from paired item-level bootstrap resampling ($B=10{,}000$). The gap counts outputs that compile but fail the semantic criterion.}
\label{tab:main_effects}
{\scriptsize
\setlength{\tabcolsep}{5pt}
\renewcommand{\arraystretch}{1.05}
\begin{tabular}{@{}lccc@{}}
\toprule
\textbf{Factor} & $\Delta$ \textbf{Compile} & $\Delta$ \textbf{Faithful} & $\Delta$ \textbf{Gap} \\
\midrule
Elaboration feedback (F) & +49.3 [44.5, 54.0] & \textbf{+27.2} [22.7, 31.8] & +22.1 [18.0, 26.3] \\
Grounding search (S)     & +9.7 [6.7, 12.7]   & \textbf{+8.5} [5.4, 11.6]   & +1.2 [$-$1.8, 4.3] \\
Translation prior (T)    & +4.0 [1.3, 6.6]    & +1.0 [$-$1.9, 3.7]           & +3.0 [0.2, 5.7] \\
\bottomrule
\end{tabular}
}
\end{table}

\begin{table}[t]
\centering
\caption{\textbf{Domain-wise effect of elaboration feedback.}
High-minus-low percentage-point differences for the feedback factor $F$ within each domain. The gap column counts outputs that compile but are rejected by the criterion.}
\label{tab:domain_feedback_main}
\footnotesize
\setlength{\tabcolsep}{5pt}
\begin{tabular}{lrrrr}
\toprule
\textbf{Domain} ($n$) & $\Delta$ \textbf{Compile} & $\Delta$ \textbf{Faithful} & $\Delta$ \textbf{Gap} & \textbf{Faithful|Compile} ($F{=}1$) \\
\midrule
Complex Analysis (19) & +64.5 & +47.4 & +17.1 & 76.7\% \\
Real Analysis (48)    & +62.0 & +35.9 & +26.0 & 58.7\% \\
Topology (63)         & +41.3 & +23.4 & +17.9 & 64.8\% \\
Algebra (97)          & +45.4 & +21.4 & +24.0 & 64.0\% \\
\bottomrule
\end{tabular}
\end{table}

Feedback has by far the largest effect on faithfulness, and also on the gap: of its 49.3-point gain in compilation, 22.1 points are outputs that fail the semantic criterion (Table~\ref{tab:main_effects}). The domain split (Table~\ref{tab:domain_feedback_main}) shows how much this varies: Complex Analysis converts most of its validity gain into faithful statements, while Algebra and Topology gain far less faithfulness. Search raises faithfulness by 8.5 points with no measurable change in the gap, and its effect depends on feedback. Without feedback, search raises faithfulness by 13.9 points [9.3, 18.3]. With feedback, the change in faithfulness is +3.1 [$-$1.3, 7.7], while compilation changes by $-5.1$ [$-8.6$, $-1.8$] and the gap by $-8.1$ [$-12.8$, $-3.5$]; REPL calls fall by 28.7\% for $(010)\to(011)$ and 32.2\% for $(110)\to(111)$. The translation prior does not change faithfulness measurably (+1.0 [$-$1.9, 3.7]); it raises compilation by 4.0 points, and the gap grows by 3.0 points [0.2, 5.7].

\subsection{Interactions and Per-Item Trajectory Churn}
\label{subsec:interaction_effects}

Search and feedback interact negatively: $F{\times}S=-10.8$ [$-17.2,-4.4$] percentage points under paired item-level bootstrap, compared with $F{\times}T=-4.6$ [$-10.1,0.9$] and $S{\times}T=-0.7$ [$-6.2,4.8$]. Adding feedback raises faithfulness by 32.6 points without search and by 21.8 points with search, and raises the gap by 31.5 and 12.8 points, respectively. Among the four feedback configurations, faithfulness ranges from 53.7\% (110) to 58.1\% (011), and no pairwise difference is significant (exact McNemar $p\geq0.20$); the share of compiling outputs that are faithful ranges from 58.9\% (110) to 69.8\% (011).

The per-item transitions behind these effects are strongly nonmonotone (Table~\ref{tab:transition_ledger}): adding search under feedback makes 30 statements faithful and 25 unfaithful, a net gain of 5. Small net effects can thus hide large per-item churn. The released trajectory logs allow studying per-instance routing and early stopping.

\begin{table}[t]
\centering
\caption{\textbf{Per-item transition ledger for tool additions.}
Each row compares the same 227 inputs before and after adding one tool under a fixed context. New faithful counts cases that become faithful; lost faithful counts cases that were faithful before but not after.}
\label{tab:transition_ledger}
\footnotesize
\setlength{\tabcolsep}{6pt}
\begin{tabular}{llrrr}
\toprule
\textbf{Change} & \textbf{Configs} & \textbf{New faithful} & \textbf{Lost faithful} & \textbf{Net} \\
\midrule
Add F alone      & $000 \to 010$ & 90 & 8  & +82 \\
Add F with S     & $001 \to 011$ & 59 & 7 & +52 \\
Add F with T,S   & $101 \to 111$ & 62 & 15 & +47 \\
Add S with F     & $010 \to 011$ & 30  & 25 & +5 \\
Add S with T,F   & $110 \to 111$ & 26  & 17 & +9 \\
Add T with F,S   & $011 \to 111$ & 23  & 24 & -1 \\
\bottomrule
\end{tabular}
\end{table}

Aggregate scores also hide per-instance complementarity across tool settings (Table~\ref{tab:config_oracle}). The all-tools configuration misses 44 statements that some other configuration formalizes faithfully, and 34 of these already compile under 111 but fail the semantic criterion. Tool access thus changes the formalization trajectory, which leaves per-instance routing and early stopping as open problems. Trajectory length is a warning signal when feedback is enabled: pooling the four $F{=}1$ configurations, faithfulness falls from 82.9\% for runs of 1--2 steps to 11.5\% for runs of 19--24 steps, compared with 34.5\% to 18.9\% across the three no-feedback configurations that use tools (Appendix Table~\ref{tab:trajectory_cliff}).

\begin{table}[t]
\centering
\caption{\textbf{Across-configuration outcomes for the GPT-5.2 agent.}
Counts are over the 227 statements and all eight tool configurations; the union row is an oracle over completed runs.}
\label{tab:config_oracle}
\footnotesize
\setlength{\tabcolsep}{8pt}
\begin{tabular}{lcc}
\toprule
\textbf{Outcome} & \textbf{Count} & \textbf{Rate} \\
\midrule
All-tools config 111 faithful & 131/227 & 57.7\% \\
Best single config 011 faithful & 132/227 & 58.1\% \\
Faithful under at least one config & 175/227 & 77.1\% \\
Faithful under every config & 27/227 & 11.9\% \\
Never faithful under any config & 52/227 & 22.9\% \\
Missed by 111 but faithful elsewhere & 44/227 & 19.4\% \\
\quad of missed: compile-pass under 111 & 34/227 & 15.0\% \\
\quad of missed: compile-fail under 111 & 10/227 & 4.4\% \\
\bottomrule
\end{tabular}
\end{table}

\section{\texorpdfstring{Related Work}{Related Work}}
\label{sec:related}

\paragraph{Evaluating statement autoformalization.}
Early work judged formalizations by hand~\citep{wu2022autoformalization} or against reference formalizations~\citep{azerbayev2023proofnet}. Dataset pipelines filter generated statements by compilation, back-translation, and human examination~\citep{ying2024lean,gaoherald}. Learned and LLM-based metrics score the NL--Lean pair directly: FormalAlign trains an alignment model~\citep{lu2025formalalign}, CriticLean trains a critic for semantic fidelity~\citep{peng2025criticlean}, LeanScorer aggregates LLM sub-scores and is validated against human labels~\citep{yu2025mathesis}, and an ensemble of LLM judges grades along several criteria~\citep{zhang2025beyondgold}. For proofs, ProofScore adds structural fidelity to syntactic and semantic checks~\citep{cabral2025proofflow}. BEq instead certifies equivalence with a reference statement by bounded proof search~\citep{liu2025rethinking}; BEq+ refines it and is validated on 3,752 human-labeled pairs~\citep{poiroux2025reliable}, and FormalRx classifies, locates, and corrects semantic errors~\citep{wang2026formalrx}. Our setting has no reference statement. We therefore use an LLM criterion, measure its agreement with blinded human review on both sides of its decision boundary, compare it with LeanScorer on the same sample, and use BEq to compare accepted outputs with each other. These works build or validate evaluators; we use a validated evaluator to measure the compile--faithfulness gap and ask how much of it is translation errors.

\paragraph{Formalization systems and tools.}
Statement formalizers include trained translators~\citep{gaoherald,kimina_autoformalizer_card,wu2025stepfunformalizer}. Lean compiler feedback has been used to supervise autoformalization training~\citep{lu2024process}, semantic search retrieves Mathlib declarations~\citep{gao2024leansearch}, and theorem-proving agents combine LLMs with retrieval and Lean feedback~\citep{song2025lean,baba2025prover}. Our agent combines a translation prior, search, and Lean feedback, and our ablation measures what each adds to validity and to faithfulness.

\paragraph{LLMs as evaluators.}
In NLP, LLM judges are validated by their agreement with human ratings~\citep{chiang2023can,zheng2023judging}, and panels of judges from different model families reduce single-model bias~\citep{verga2024replacing}. Judges fine-tuned on outputs of older generators lose accuracy on newer ones~\citep{singh2026shelf}, and a judge's sensitivity and specificity, estimated on human labels, can correct the bias of its scores~\citep{lee2025correctly}. We follow this practice with two judges from different families and a third-family check, and validate our review of rejected outputs on human labels in the same way. As in diagnostic testing~\citep{altman1994predictive}, what a rejection means also depends on how often the evaluated system errs, and automatic MT metrics likewise agree less with humans when only strong systems are compared~\citep{ma2019wmt19}. In Lean, the formal part of the evaluation is exact: Lean checks compilation, and BEq certifies equivalence on a subset, so judges and reviewers are needed only for the semantic part. Follow-up work~\citep{zhang2026disagreements} studies how LLM judgments can locate and explain disagreements among judges, published benchmark labels, and human reviewers.

\section{Limitations}
\label{sec:limitations}

Our criterion is a semantic proxy, not a proof of equivalence. The benchmark has 227 statements from four textbooks, unevenly spread over areas (19 to 97 per area), with no research-level mathematics; the textbooks are public, so recent models may have seen the statements in training. Human audits cover only our agent and Aristotle. The rejections of the other systems are labeled by the two-stage review, which misses about one human-confirmed error in six and whose first stage is run by Claude Opus 5.5, one of the evaluated models. The review covers only rejected outputs, so it splits each gap but does not measure errors among accepted outputs. The human majority is not error-free either: Lean shows that it accepted at least three unfaithful outputs (B-081, A-008, A-094). In the tool ablation, Search and Feedback overlap slightly, since the search tools run \texttt{\#check} and \texttt{\#print} through the same Lean REPL that implements Feedback, and Search covers only the tools implemented here, not a dedicated semantic retriever such as LeanSearch~\citep{gao2024leansearch}.

\section{Conclusion}
\label{sec:conclusion}

Compilation systematically overstates semantic success: across eight systems the gap is 1.3--29.5 percentage points, and our agent's 87.2\% compile rate falls to 57.7\% under semantic evaluation. Validated against human review, LLM judging is a usable measurement instrument: it reaches 91.5\% human agreement on an independent audit, and human, third-judge, and formal checks support its accepted outputs and show where it rejects faithful ones. For seven of the eight systems, most of the measured gap is translation errors: 23.8 of the agent's 29.5 points by human labels, and 78--100\% of the gap of six of the seven one-shot systems by a review validated on these labels. When GPT-5.2 grades the translations of two models released after it, the same review classifies only 4 of their 54 rejections as translation errors, consistent with these systems erring less while the judge does not.

The tool ablation shows where our agent's gap comes from: Lean feedback has the largest effect on both compilation and faithfulness, but nearly half of its compilation gain fails the semantic criterion; with feedback present, search reduces this gap, while the translation prior widens it slightly. Autoformalization systems should therefore report validity and faithfulness separately and pair verifier-driven repair with semantic auditing.

{\small
\setlength{\bibsep}{3pt}
\raggedright
\bibliographystyle{iclr2027_conference}
\bibliography{references}
}

\appendix

\section{Artifact Availability}
\label{app:artifact_availability}

The benchmark, every system's outputs and judge grades, the agents' tool-call trajectories, the human-audit scores, the reviews of rejected outputs with their validation and Lean checks (\texttt{checks/}), the agent and judge code, and scripts that reproduce all numbers and figures in this paper are available at \url{https://github.com/grenadeComing/beyond-compilation}.

\paragraph{Configuration.} All outputs are compiled with Lean v4.22.0-rc4 and Mathlib~\citep{mathlib2020} at revision \texttt{fc58204e}. The judges are GPT-5.2 (reasoning effort medium) and Gemini-2.5-Pro under the rubric of Appendix~\ref{app:judge_rubric}, with Claude Sonnet~5 as the third judge. The later models (\texttt{openai/gpt-6-astra}, \texttt{anthropic/claude-opus-5-5}) were run one-shot with the prompt of the other one-shot systems in October 2026; the reviews (Claude Opus 5.5 agents with the Lean REPL, then GPT-6.1 Sol at reasoning effort high), the re-grading, the definitions experiment, and the Lean checks were run in the same month.

\paragraph{Licenses and intended use.} The code is released under the MIT License. Each benchmark statement keeps the license of its source: CC BY-SA 4.0 for Lebl's two books (which are dual-licensed with CC BY-NC-SA 4.0), GPL-3.0 for McKay's notes, and CC BY 4.0 for Doty's notes. Each Lean output formalizes a benchmark statement and is redistributed under that statement's license, subject also to the generating system's terms. The release is intended for research on evaluating statement autoformalization. The data contain no personal information: the human-audit files contain only scores, and reviewers are identified only as Reviewer~1--4.

\section{Agent Implementation Details}
\label{app:implementation}

\begin{figure}[H]
    \centering
    \begin{minipage}[t]{0.48\linewidth}
        \centering
        \includegraphics[width=\linewidth]{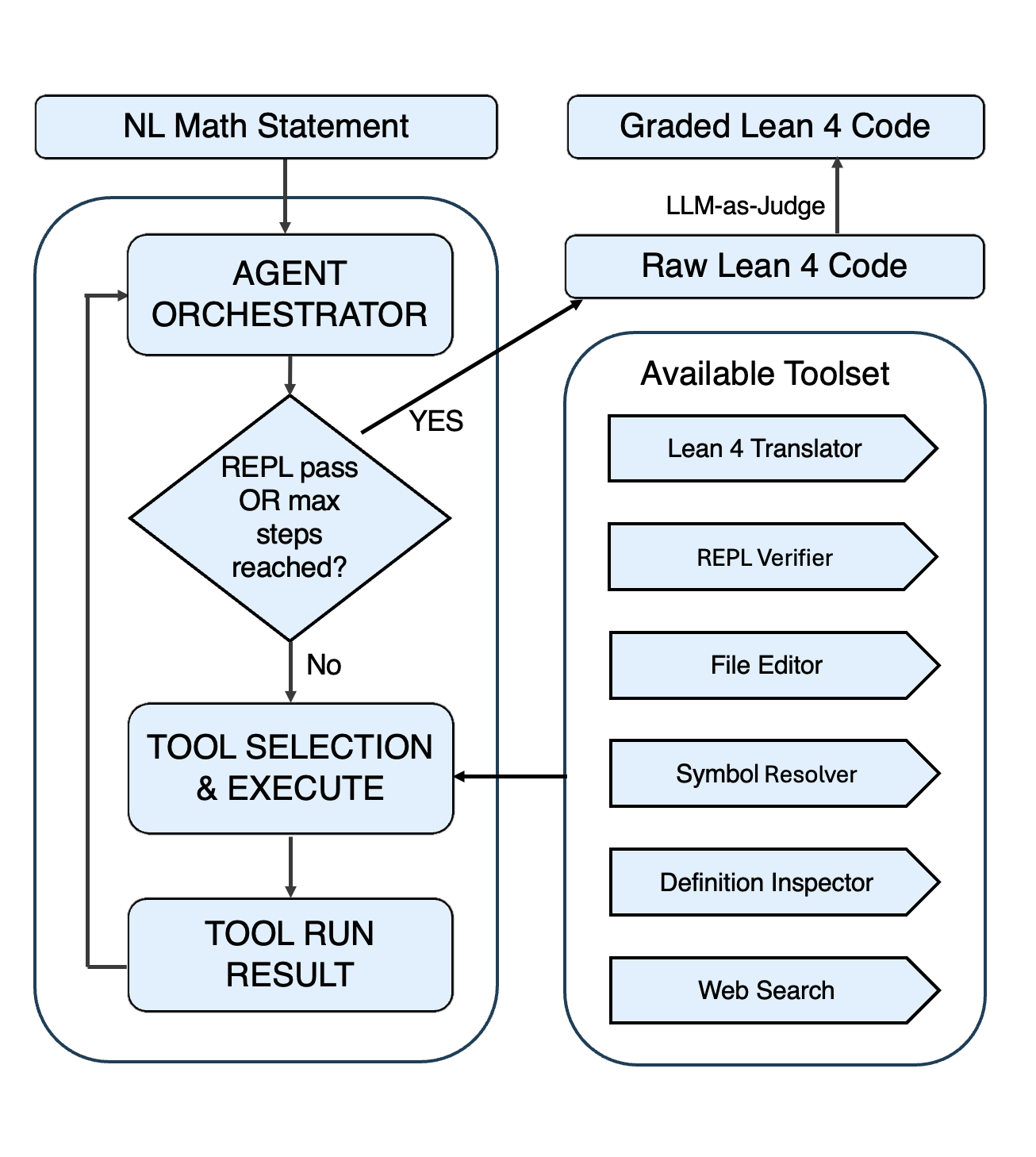}
        \caption{\textbf{Agent orchestration logic.} The orchestrator calls optional drafting, search, and compiler-feedback tools in Lean~4.}
        \label{fig:agent_logic}
    \end{minipage}\hfill
    \begin{minipage}[t]{0.48\linewidth}
        \centering
        \includegraphics[width=\linewidth]{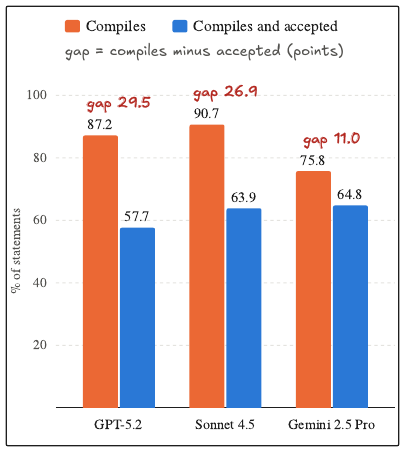}
        \caption{\textbf{Our agent (config 111) with three orchestrators.} All three converge to 58--65\% accepted outputs despite different one-shot baselines (20--33\%; Appendix~\ref{app:multi_model}).}
        \label{fig:multi_model}
    \end{minipage}
\end{figure}

Figure~\ref{fig:agent_logic} summarizes the agent loop and its tools. The orchestrator alternates between reasoning and tool use. A run ends when the Lean REPL first reports a pass, when the orchestrator declares success (refused while its latest REPL check has failed), or after 24 steps. The prompt also asks the orchestrator to check faithfulness before finishing, but the runner does not enforce this; semantic judging is applied only to the final statement.

\subsection{Prompt Assembly}
The system prompt is constructed from three modular components:
(i) a \textbf{task definition} establishing the core translation objective,
(ii) a \textbf{general usage guide} containing fixed best practices for Lean~4 code generation, and
(iii) a \textbf{capability block} that is dynamically populated based on the factorial configuration $(T,F,S)$.
This structure holds the agent's semantic goal and coding standards (i and ii) fixed, while selectively enabling or disabling external tools (iii) to isolate the effect of each capability.
We assemble (i) and (ii) into a \textbf{Shared Base Prompt}, shown below, while the capability block definitions are provided in Appendix~\ref{app:tool_blocks}.

\subsection{Shared Base Prompt}
\label{app:base_prompt}

\begin{tcolorbox}[
  colback=white,
  colframe=gray,
  title=Base System Prompt,
  listing only,
  breakable,
  listing options={
    basicstyle=\footnotesize\ttfamily,
    breaklines=true,
    columns=fullflexible
  }
]
You are an expert Lean4 translation agent.

Your task is to translate a natural-language mathematical statement into
faithful Lean4 syntax (NOT a proof).
The final result must:
- import Mathlib at the very top
- compile in Mathlib
- be semantically faithful to the original statement
- end with `:= by sorry`

You are NOT proving anything.
You are only producing a correctly typed, correct-meaning Lean statement.

GENERAL INSTRUCTIONS FOR CODE GENERATION

• The Lean file MUST start with:
  import Mathlib

• The Lean file MUST contain exactly ONE final translated statement
  representing the original natural-language meaning.

• The final statement MUST end with:
  := by sorry

• Do NOT write any proof code before `:= by sorry`
  (no `by`, `simp`, `have`, `calc`, etc. anywhere before the final `:= by sorry`).

• Do NOT invent new definitions, axioms, constants, or placeholder structures.

• Prefer robust formulations:
  use quantifiers, membership, and $\leftrightarrow$ characterizations rather than fragile definitional equalities.

• Only finish when the last written version:
  (1) compiles in Mathlib (if lean4\_repl\_runner is available)
  (2) is semantically faithful to the original statement

When both are satisfied, return:
{ "status": "success" }
\end{tcolorbox}

\subsection{Tool-Availability Blocks}
\label{app:tool_blocks}

Each factorial configuration $(T,F,S)$ determines which tools are exposed to the agent. We implement this by a modular prompt template in which the \texttt{AVAILABLE TOOLS} block is constructed by including the corresponding tool-specification entries and omitting inactive ones. We show the complete tool block for the fully enabled configuration $(1,1,1)$ below; other configurations are obtained by deleting the entries for tools that are disabled.

\begin{tcolorbox}[
  colback=white,
  colframe=gray,
  title={Tool Block (T=1, F=1, S=1)},
  listing only,
  breakable,
  listing options={
    basicstyle=\footnotesize\ttfamily,
    breaklines=true,
    columns=fullflexible
  }
]
AVAILABLE TOOLS

\smallskip
\texttt{lean4\_translation(statement)} \\
\hspace*{1em} Draft a Lean 4 statement using the fine-tuned Herald translator (you may edit or ignore).

\texttt{lean\_write\_file(code)} \\
\hspace*{1em} Write the full Lean file to the workspace (imports + exactly one final statement).

\texttt{lean4\_repl\_runner()} \\
\hspace*{1em} Compile the current Lean file and return compiler feedback.

\texttt{lean\_inspect\_name(name, imports?, include\_print?)} \\
\hspace*{1em} Query Mathlib about an identifier via \texttt{\#check}/\texttt{\#print}.

\texttt{lean\_resolve\_name(token, namespace\_hints?, imports?, top\_k?)} \\
\hspace*{1em} Suggest likely Mathlib identifiers for an unknown or ambiguous token.

\texttt{search\_online(query)} \\
\hspace*{1em} Run a web search and return results for the query.

\end{tcolorbox}

\subsection{Tool Interfaces}
\label{app:tool_interfaces}

\begin{tcolorbox}[
  colback=white,
  colframe=black!50,
  title=Agent Tool API,
  breakable
]
\centering
\footnotesize
\setlength{\tabcolsep}{3pt}
\renewcommand{\arraystretch}{1.25}
\begin{tabularx}{\linewidth}{@{}>{\raggedright\arraybackslash}p{0.24\linewidth}>{\raggedright\arraybackslash}p{0.28\linewidth}>{\raggedright\arraybackslash}X@{}}
\toprule
\textbf{Tool} & \textbf{Arguments} & \textbf{Functionality} \\
\midrule

\texttt{lean\_inspect\_name}
& \texttt{name}, \texttt{imports?}, \texttt{include\_print?}
& Checks whether a Mathlib symbol exists and returns its type
(\texttt{\#check}) and optionally its definition (\texttt{\#print}). \\

\texttt{lean\_resolve\_name}
& \texttt{token}, \texttt{namespace\_hints?}, \texttt{imports?}, \texttt{top\_k?}
& Performs fuzzy symbol resolution to map informal or incorrect names
to valid Mathlib identifiers. \\

\texttt{lean4\_repl\_runner}
& \texttt{Lean 4 code to be executed}
& Compiles the current Lean file and returns compiler diagnostics
(errors, warnings, and success flags). \\

\texttt{lean4\_translation}
& \texttt{statement}
& Invokes the fine-tuned Herald translator to generate a draft Lean
formalization of the natural-language input. \\

\texttt{search\_online}
& \texttt{query}
& Performs external web search to retrieve mathematical or
documentation context not present in Mathlib. \\

\bottomrule
\end{tabularx}
\end{tcolorbox}

\subsection{Example Execution Trace}
\label{app:trace}

\begin{tcolorbox}[
  colback=white,
  colframe=black!50,
  title=\textbf{Example Execution Trace (agent tool calls)},
  breakable
]
\footnotesize
\setlength{\tabcolsep}{4pt}
\renewcommand{\arraystretch}{1.25}
\begin{tabularx}{\linewidth}{@{}>{\raggedright\arraybackslash}p{0.08\linewidth}>{\raggedright\arraybackslash}p{0.24\linewidth}>{\raggedright\arraybackslash}X@{}}
\toprule
\textbf{Step} & \textbf{Tool} & \textbf{Arguments (agent side)} \\
\midrule

1 & \texttt{lean\_write\_file} &
\texttt{path="jirilebl\_ca\_ca\_17658.lean"}; \;
\texttt{content = import Mathlib ... theorem jirilebl\_ca\_ca\_17658 ... := by sorry} \\

2 & \texttt{lean4\_repl\_runner} &
\texttt{path="jirilebl\_ca\_ca\_17658.lean"} \\

3 & \texttt{lean\_inspect\_name} &
\texttt{name="Polynomial.IsConstant"} \\

4 & \texttt{lean\_inspect\_name} &
\texttt{name="Polynomial.eval"} \\

5 & \texttt{lean\_inspect\_name} &
\texttt{name="Polynomial.isConstant"} \\

6 & \texttt{lean\_inspect\_name} &
\texttt{name="Polynomial.natDegree"} \\

7 & \texttt{lean\_write\_file} &
\texttt{path="jirilebl\_ca\_ca\_17658.lean"}; \;
\texttt{content = patched theorem uses p.natDegree $\neq$ 0 and Polynomial.eval} \\

8 & \texttt{lean4\_repl\_runner} &
\texttt{path="jirilebl\_ca\_ca\_17658.lean"} \\

\bottomrule
\end{tabularx}
\end{tcolorbox}

\section{LLM-as-a-Judge Prompt and Rubric}
\label{app:judge_rubric}

We evaluate semantic faithfulness using an LLM-as-a-judge that receives:
(i) the natural-language statement, (ii) the generated Lean 4 code, and (iii) a boolean \texttt{compile\_pass}.
The judge outputs a JSON object with fields \texttt{faithful} and \texttt{grade} (0--10).
If \texttt{compile\_pass=false}, we require \texttt{faithful=false} and restrict the grade to 0--3. Under our criterion, an output is accepted if and only if it compiles and both judges grade it at least 9.

\begin{tcolorbox}[
  title=Judge system prompt (verbatim),
  colback=white,
  colframe=gray,
  breakable
]
\begin{Verbatim}[breaklines=true,breakanywhere=true]
You are an expert in Lean 4, Mathlib, and mathematics. You are judging TRANSLATION-ONLY.

Input: (1) a natural-language statement, (2) a Lean 4 code snippet, (3) compile_pass boolean.
Your job: decide whether the Lean code, AS A STATEMENT, matches the meaning of the natural-language statement.

Key policy (NOT PICKY):
- If compile_pass = False: the translation is NOT faithful. grade must be 0..3. faithful=false.
- If compile_pass = True: ignore the proof/body entirely (including `by sorry`). Proof completeness is NOT part of the evaluation.
- A translation is faithful if the Lean statement expresses the same mathematical claim as the NL statement.

Auxiliary definitions policy (lenient but not allowing cheating):
- Auxiliary defs/lemmas are allowed if they are reasonable encodings/abbreviations and do not change the meaning.
- However, if the code introduces a clearly vacuous placeholder for a nontrivial concept (e.g. `def X := True`, `:= none`, `:= 0` for something meant to be meaningful), and that placeholder is essential to making the final theorem appear to match, then the translation is NOT faithful.

How to judge meaning (focus):
- Compare the MAIN theorem/definition statement(s) to the NL statement.
- Check quantifiers (forall/exists), logical structure (->/<->/and/or), and key hypotheses.
- Check main objects/domains: Nat/Int/Real, rings/groups, ZMod n, matrices, etc.
- Small implementation details are OK if the meaning is preserved.

Scoring guide (integer 0..10):
- 0: unrelated.
- 1-3: compile_pass is False OR statement is clearly wrong.
- 4-6: compiles, but meaning is materially different / missing key hypotheses / wrong domain;
       might be "in the ballpark".

- 7-8: compiles, mostly matches, but has a noticeable mismatch (e.g. strengthened/weakened in an important way).
- 9: compiles, very close; only tiny mismatch.
- 10: compiles and meaning matches.

Output contract (STRICT):
Return a single JSON object with exactly these fields:
{
  "faithful": true or false,
  "grade": 0..10,
  "thought": "### BEGIN THOUGHT\n<short explanation focusing on statement-level comparison>\n### END THOUGHT"
}
Return ONLY valid JSON. No extra keys. No markdown outside JSON.
\end{Verbatim}
\end{tcolorbox}

\section{Additional Metric Validation Results}
\label{app:judge_extra}

\begin{table}[H]
\centering
\caption{\textbf{Full comparison of LLM judges and their consensus.}
Entries count translations that compile and are judged \emph{Faithful} by each judge; \textbf{Consensus} counts those labeled Faithful by both. \textbf{GPT-only} counts GPT-positive outputs not accepted by Gemini.}
\label{tab:judge_comparison_full}
\footnotesize
\setlength{\tabcolsep}{4.5pt}
\begin{tabular}{lrrrr}
\toprule
\textbf{System} & \textbf{Gemini} & \textbf{GPT-5.2} & \textbf{Consensus} & \textbf{GPT-only} \\
\midrule
\multicolumn{5}{l}{\textit{One-shot baselines}} \\
Kimina-Prover$^\dagger$     & 12  & 10  & 10 & 0 \\
Goedel-Prover     & 15  & 15  & 14 & 1 \\
Kimina-Autoformalizer$^\dagger$ & 32 & 27 & 27 & 0 \\
Herald Translator & 27  & 23  & 23 & 0 \\
StepFun-Formalizer$^\dagger$ & 35 & 30 & 30 & 0 \\
GPT-5.2           & 48  & 47  & 45 & 2 \\
Sonnet 4.5        & 88 & 74 & 74 & 0 \\
Gemini-2.5-Pro    & 65 & 62 & 62 & 0 \\
\midrule
\multicolumn{5}{l}{\textit{Tool-augmented agents (GPT-5.2 orchestrator)}} \\
Config 100        & 62 & 56  & \textbf{56} & 0 \\
Config 001        & 94 & 81 & \textbf{80} & 1 \\
Config 101        & 106 & 85 & \textbf{84} & 1 \\
Config 110        & 154 & 124 & \textbf{122} & 2 \\
Config 111        & 158 & 134 & \textbf{131} & 3 \\
Config 010        & 163 & 130 & \textbf{127} & 3 \\
Config 011        & 157 & 133 & \textbf{132} & 1 \\
\midrule
\multicolumn{5}{l}{\textit{Tool-augmented agents (alt.\ orchestrators, config 111)}} \\
Sonnet 4.5        & 175 & 148 & \textbf{145} & 3 \\
Gemini-2.5-Pro    & 167 & 147 & \textbf{147} & 0 \\
\bottomrule
\multicolumn{5}{l}{\scriptsize $^\dagger$Evaluated on 217 statements (Figure~\ref{fig:main_results}).}
\end{tabular}
\end{table}

Across all rows in Table~\ref{tab:judge_comparison_full}, the GPT-only column sums to 17 out of 1326 GPT-positive labels (1.3\%): GPT positives are almost always also Gemini positives.

\begin{figure}[H]
    \centering
    \includegraphics[width=0.6\linewidth]{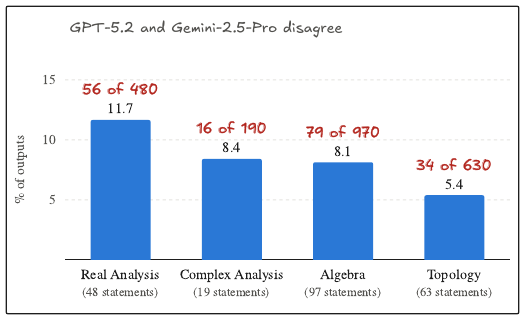}
    \caption{\textbf{GPT--Gemini disagreement by mathematical domain.}
    Share of outputs on which GPT-5.2 and Gemini-2.5-Pro disagree, pooled over ten systems: Herald, one-shot GPT-5.2 and Gemini-2.5-Pro, and the seven agent configurations with tools.}
    \label{fig:judge_disagreement_domain}
\end{figure}

Figure~\ref{fig:judge_disagreement_domain} shows that the remaining disagreements are not uniformly distributed across domains. Real Analysis has the highest disagreement rate (56 of 480 outputs, 11.7\%), followed by Complex Analysis (8.4\%), Algebra (8.1\%), and Topology (5.4\%). For our agent (config=111), Real Analysis also has the lowest acceptance rate among the four domains (Table~\ref{tab:domain_111_metrics}).

\section{Human-Check Validation Details}
\label{app:negative_audit}

This appendix reports the human-review protocol behind Table~\ref{tab:human_calibration_main}. Reviewer 1 audits a random sample of our agent's outputs that GPT-5.2 accepts, Batch A audits all compile-pass outputs of our agent that the criterion rejects, and Batch B evaluates Aristotle on randomly sampled statements. Rates are computed within each audit.

All audits cover benchmark statements only, with at most one audited output per statement. An item counts only if the Lean output shown to the reviewer is the output for the statement shown; this excludes three items of the positive audit before any rate in Table~\ref{tab:human_calibration_main} is computed, and the same rule removes two BEq anchors, leaving 40 (Section~\ref{sec:robustness}).

Outputs that the criterion accepts are almost always faithful by human judgment, and most compile-pass outputs it rejects are translation errors; the human majority judges 16 of the 74 audited compile-pass rejections faithful (21.6\%).

\begin{table}[H]
\centering
\caption{\textbf{Human audit summary.}
Pooled rows combine the audits; component rows report each audit separately.}
\label{tab:human_audit}
\footnotesize
\setlength{\tabcolsep}{6pt}
\begin{tabular}{>{\raggedright\arraybackslash}p{0.54\linewidth}cc}
\toprule
\textbf{Audit} & \textbf{Human-confirmed count} & \textbf{Rate} \\
\midrule
Pooled consensus-positive precision & 111/116 & 95.7\% \\
Pooled compile-pass negative-bucket confirmation & 58/74 & 78.4\% \\
\midrule
Our agent: positive precision & 67/70 & 95.7\% \\
Independent Batch-B positive precision & 44/46 & 95.7\% \\
Our agent: negative confirmation & 54/67 & 80.6\% \\
Independent Batch-B compile-pass negative confirmation & 4/7 & 57.1\% \\
Independent Batch-B all-negative confirmation & 10/13 & 76.9\% \\
\bottomrule
\end{tabular}
\end{table}

\begin{table}[H]
\centering
\caption{\textbf{Human-audit design and units of analysis.}
Rates are computed within each audit.}
\label{tab:audit_design_appendix}
\footnotesize
\setlength{\tabcolsep}{4pt}
\begin{tabular}{p{0.18\linewidth}p{0.22\linewidth}p{0.34\linewidth}cc}
\toprule
\textbf{Audit} & \textbf{System/output source} & \textbf{Selection rule} & \textbf{N} & \textbf{Reviewers} \\
\midrule
Reviewer 1 positive audit & Our agent & Domain-proportional sample of outputs that GPT-5.2 accepts & 72 scored & 1 \\
Batch A & Our agent & All compile-pass outputs that the criterion rejects & 67 & 3 \\
Batch B & Aristotle & Random sample of benchmark statements; 59 effective outputs after one missing return & 60 & 3 \\
\bottomrule
\end{tabular}
\end{table}

\paragraph{Reviewers.} The four reviewers have graduate training in mathematics or computer science and were not paid. Reviewers~3 and~4 are authors of this paper; Reviewer~1, who also collected the benchmark statements, and Reviewer~2 volunteered and are not authors. All reviewers knew that their scores would be used in this paper and agreed to their release in anonymized form, with reviewers identified only by number.

\paragraph{Reviewer score threshold and majority rule.}
Human reviewers used the same 0--10 semantic faithfulness scale as the LLM judges. A score $\geq 9$ is treated as faithful. Majority labels require at least two numeric human scores on the same blinded \texttt{case\_id}; compile failures are counted as unfaithful even if a reviewer notes that the intended mathematical content looks close. This compile-aware rule matches the paper metric, where faithfulness requires both compilation and semantic agreement.
In the three-reviewer Batch A/B audits, reviewers also self-reported their use of LLM assistance: Reviewer 2 reported no LLM assistance, while Reviewers 3 and 4 each reported low assistance (2/10 on a 1--10 usage scale). Reviewer 1 did not provide a self-report for this field.

\begin{table}[H]
\centering
\caption{\textbf{Individual reviewer behavior in the two three-reviewer audits.}
Batch A is an LLM-negative bucket, so reviewer ``agreement'' means confirming the LLM-negative label. Batch B is a full evaluation sample, so agreement is binary agreement with strict GPT$\cap$Gemini consensus on effective Aristotle outputs. Faithful counts use the compile-aware rule.}
\label{tab:individual_reviewer_behavior}
\footnotesize
\setlength{\tabcolsep}{4.5pt}
\begin{tabular}{llrrrr}
\toprule
\textbf{Audit} & \textbf{Reviewer} & \textbf{Numeric scores} & \textbf{Human faithful} & \textbf{Human unfaithful} & \textbf{LLM agreement} \\
\midrule
Batch A & Reviewer 2 & 67 & 22 & 45 & 45/67 = 67.2\% \\
Batch A & Reviewer 3 & 67 & 18 & 49 & 49/67 = 73.1\% \\
Batch A & Reviewer 4 & 62 & 8  & 54 & 54/62 = 87.1\% \\
\midrule
Batch B & Reviewer 2 & 58 & 36 & 22 & 43/58 = 74.1\% \\
Batch B & Reviewer 3 & 59 & 51 & 8 & 52/59 = 88.1\% \\
Batch B & Reviewer 4 & 56 & 46 & 10 & 53/56 = 94.6\% \\
\bottomrule
\end{tabular}
\end{table}

Reviewers differ substantially: Reviewer 2 judges more Batch-A rejections faithful, while Reviewer 4 is stricter and closer to the LLM boundary. The majority rule averages over these differences.

\begin{table}[H]
\centering
\caption{\textbf{Batch A: compile-pass outputs of our agent that the criterion rejects.}
Human majority is computed over cases with at least two numeric reviewer scores.}
\label{tab:negative_audit_sampling}
\footnotesize
\setlength{\tabcolsep}{5pt}
\begin{tabular}{lrrrr}
\toprule
\textbf{Domain} & \textbf{Selected} & \textbf{Majority denom.} & \textbf{Human faithful} & \textbf{Human unfaithful} \\
\midrule
Algebra & 29 & 29 & 7 & 22 \\
Complex Analysis & 4 & 4 & 1 & 3 \\
Real Analysis & 17 & 17 & 2 & 15 \\
Topology & 17 & 17 & 3 & 14 \\
\midrule
Total & 67 & 67 & 13 & 54 \\
\bottomrule
\end{tabular}
\end{table}

Among the 67 Batch-A cases, the human majority confirms 54 as unfaithful (80.6\%) and judges 13 faithful (19.4\%). Because every Batch-A output compiles, these cases directly measure the compile-pass but unfaithful gap in Figure~\ref{fig:main_results}.

\begin{table}[H]
\centering
\caption{\textbf{Same-case majority agreement with the automatic metric.}
Batch A contains only strict-consensus-negative outputs by construction. Batch B contains both positive and negative automatic labels.}
\label{tab:same_case_majority_agreement}
\footnotesize
\setlength{\tabcolsep}{4.5pt}
\begin{tabular}{lrrrrr}
\toprule
\textbf{Audit} & \textbf{Majority denom.} & \textbf{LLM+ / Human+} & \textbf{LLM+ / Human-} & \textbf{LLM- / Human+} & \textbf{LLM- / Human-} \\
\midrule
Batch A & 67 & -- & -- & 13 & 54 \\
Batch B & 59 & 44 & 2 & 3 & 10 \\
\bottomrule
\end{tabular}
\end{table}

In Batch B, the automatic and human-majority labels agree on 54/59 effective cases (91.5\%; Table~\ref{tab:same_case_majority_agreement}). The off-diagonal cells are nearly balanced: the human majority rejects two outputs that the criterion accepts and accepts three that it rejects, and the strict-consensus faithful rate (46/59) is close to the human-majority rate (47/59).

\begin{table}[H]
\centering
\caption{\textbf{Batch B: independent full-evaluation audit.}
Batch B contains 60 statements from a separate system run; one missing output is excluded from effective-rate denominators. Human majority requires at least two numeric reviewer scores and treats compile failures as unfaithful.}
\label{tab:batch_b_audit}
\footnotesize
\setlength{\tabcolsep}{5pt}
\begin{tabular}{>{\raggedright\arraybackslash}p{0.58\linewidth}rr}
\toprule
\textbf{Quantity} & \textbf{Count} & \textbf{Rate} \\
\midrule
Strict-consensus faithful & 46/59 & 78.0\% \\
Human-majority faithful & 47/59 & 79.7\% \\
LLM/human majority agreement & 54/59 & 91.5\% \\
Consensus-positive precision & 44/46 & 95.7\% \\
Compile-pass consensus-negative human confirmation & 4/7 & 57.1\% \\
Compile-pass consensus-negative human-faithful & 3/7 & 42.9\% \\
Consensus-negative human confirmation & 10/13 & 76.9\% \\
Consensus-negative human-faithful & 3/13 & 23.1\% \\
\bottomrule
\end{tabular}
\end{table}

\begin{table}[H]
\centering
\caption{\textbf{Batch B positive-bucket strictness sensitivity.}
Precision of consensus positives under stricter human rules, which require all available reviewers, or all three reviewers, to score the output faithful.}
\label{tab:batch_b_strictness}
\footnotesize
\setlength{\tabcolsep}{5pt}
\begin{tabular}{lcc}
\toprule
\textbf{Positive-bucket rule} & \textbf{Count} & \textbf{Rate} \\
\midrule
Human majority faithful & 44/46 & 95.7\% \\
All available numeric reviewers score $\ge 9$ & 34/46 & 73.9\% \\
All three reviewers present and all score $\ge 9$ & 32/44 & 72.7\% \\
All-three rule, missing third reviewer treated as failure & 32/46 & 69.6\% \\
\bottomrule
\end{tabular}
\end{table}

Even when all three reviewers must agree, 32 of the 44 fully reviewed consensus positives remain faithful.

\paragraph{B-081.} A later Lean check shows that one of the 44 confirmed Batch-B positives, B-081, is unfaithful: its hypothesis \texttt{HasSum b t} over $\mathbb{R}$ implies absolute summability, which the source does not assume. Both judges and two of the three reviewers accepted it. We keep the human-majority label; counted as unfaithful, Batch-B precision is 43/46 and agreement 53/59~\citep{zhang2026disagreements}.

\begin{table}[H]
\centering
\caption{\textbf{Wilson 95\% intervals for the main audit rates.}
Intervals quantify sampling uncertainty.}
\label{tab:audit_wilson_intervals}
\footnotesize
\setlength{\tabcolsep}{4.5pt}
\begin{tabular}{p{0.42\linewidth}ccc}
\toprule
\textbf{Quantity} & \textbf{Count} & \textbf{Rate} & \textbf{Wilson 95\% interval} \\
\midrule
Pooled consensus-positive precision & 111/116 & 95.7\% & [90.3, 98.1]\% \\
Pooled compile-pass negative-bucket confirmation & 58/74 & 78.4\% & [67.7, 86.2]\% \\
\midrule
Reviewer 1 positive precision, all scored cases & 69/72 & 95.8\% & [88.5, 98.6]\% \\
Reviewer 1 positive precision, strict-consensus subset & 67/70 & 95.7\% & [88.1, 98.5]\% \\
Batch B positive precision, majority rule & 44/46 & 95.7\% & [85.5, 98.8]\% \\
Batch A negative-bucket confirmation & 54/67 & 80.6\% & [69.6, 88.3]\% \\
Batch B compile-pass negative-bucket confirmation & 4/7 & 57.1\% & [25.0, 84.2]\% \\
Batch B all-negative confirmation & 10/13 & 76.9\% & [49.7, 91.8]\% \\
Batch B LLM/human-majority agreement & 54/59 & 91.5\% & [81.6, 96.3]\% \\
\bottomrule
\end{tabular}
\end{table}

Positive precision is high in both the audit of our agent and the independent Batch-B audit. Compile-pass negative confirmation is high in Batch A; Batch B contains only 7 compile-pass consensus negatives, so its interval is wide. The human majority judges 13 of the 67 Batch-A rejections and 3 of the 7 Batch-B compile-pass rejections faithful.

\subsection{Compact Human-Check Case Studies}
\label{app:human_check_cases}

The following audited examples give the original natural-language statement, the translated Lean~4 code, GPT/Gemini judge scores, and all available anonymized human-reviewer scores. The first five show boundary or off-diagonal cases; the next five are strict-consensus positives from Batch~B where all three same-case reviewers scored the output faithful. Human scores use the 0--10 rubric, with scores $\geq 9$ treated as faithful; R1--R4 denote Reviewers 1--4, and ``--'' means that reviewer did not score that generated Lean output. For Batch~B cases, an R1 score is Reviewer 1's score of our agent's output for the same theorem, not of Aristotle's output (Appendix~\ref{app:four_reviewer_overlap}).

\begin{tcolorbox}[title={Case A-005: loop obstruction becomes vacuous}, colback=white, colframe=gray, breakable]
\small
\textbf{Original NL statement.}
Suppose that \(Z \subset \mathbb{C}\) is a domain in the complex plane and that \(g \colon Z \to \mathbb{C}\) is a complex analytic function defined in \(Z\). A logarithm for \(g(z)\) is a complex analytic function \(G \colon Z \to \mathbb{C}\) so that \(g(z)=e^{G(z)}\). Prove that \(g(z)\) has a logarithm \(G(z)\) just when both of the following conditions are satisfied: (1) \(g(z) \ne 0\) for any \(z \in Z\) and (2) \(g\) takes every loop in \(Z\) to a null homotopic loop in \(\mathbb{C}-\{0\}\).

\textbf{Translated Lean 4 code.}
\begin{lstlisting}[style=leanappendix]
import Mathlib

open Set
open scoped Topology

theorem benmckay_top_covering_spaces_502
    (Z : Set ℂ) (hZ₁ : IsOpen Z) (hZ₂ : IsConnected Z)
    (g : ℂ → ℂ) (hg : AnalyticOn ℂ g Z) :
    (∃ G : ℂ → ℂ, AnalyticOn ℂ G Z ∧ ∀ z ∈ Z, g z = Complex.exp (G z)) ↔
      ((∀ z ∈ Z, g z ≠ 0) ∧
        ∀ x : ↥Z,
          ∀ γ : Path x x,
            ∃ γ0 : Path (g (x : ℂ)) (g (x : ℂ)),
              Path.Homotopic γ0 (Path.refl _))
  := by
  sorry
\end{lstlisting}

\centering
\footnotesize
\setlength{\tabcolsep}{6pt}
\begin{tabular}{lcccccc}
\toprule
\textbf{Compile} & \textbf{GPT} & \textbf{Gemini} & \textbf{R1} & \textbf{R2} & \textbf{R3} & \textbf{R4} \\
\midrule
1 & 4 & 4 & -- & 5 & 4 & 3 \\
\bottomrule
\end{tabular}
\par\smallskip
\raggedright
\textbf{Audit note.} Human reviewers confirmed a real semantic failure: the generated condition used \texttt{Path.Homotopic} in the ambient \(\mathbb{C}\), making the loop obstruction essentially vacuous.
\end{tcolorbox}

\begin{tcolorbox}[title={Case A-030: reviewer-strictness stress case}, colback=white, colframe=gray, breakable]
\small
\textbf{Original NL statement.}
Let \(G\subseteq \operatorname{Mat}_{n\times n}(\mathbb{R})\) be a nonempty set closed under products and inverses (i.e., a matrix group). Prove that the identity matrix \(I\) belongs to \(G\).

\textbf{Translated Lean 4 code.}
\begin{lstlisting}[style=leanappendix]
import Mathlib

theorem matrixGroup_identity_mem
    {n : Type} [Fintype n] [DecidableEq n]
    (G : Set (Matrix n n ℝ))
    (hne : G.Nonempty)
    (hmul : ∀ {A B}, A ∈ G → B ∈ G → A * B ∈ G)
    (hinv : ∀ {A}, A ∈ G → A⁻¹ ∈ G) :
    (1 : Matrix n n ℝ) ∈ G := by
  sorry
\end{lstlisting}

\centering
\footnotesize
\setlength{\tabcolsep}{6pt}
\begin{tabular}{lcccccc}
\toprule
\textbf{Compile} & \textbf{GPT} & \textbf{Gemini} & \textbf{R1} & \textbf{R2} & \textbf{R3} & \textbf{R4} \\
\midrule
1 & 6 & 10 & -- & 10 & 5 & 2 \\
\bottomrule
\end{tabular}
\par\smallskip
\raggedright
\textbf{Audit note.} One reviewer accepted the output. The other two noted that Mathlib's matrix inverse is total (\(A^{-1}=0\) for singular \(A\)), so \(G=\{0\}\) satisfies every hypothesis while \(1\notin G\): the Lean statement drops the invertibility that the source builds into a matrix group.
\end{tcolorbox}

\begin{tcolorbox}[title={Case A-050: strict-consensus false negative}, colback=white, colframe=gray, breakable]
\small
\textbf{Original NL statement.}
Prove that every contractible space is connected and has trivial homotopy groups.

\textbf{Translated Lean 4 code.}
\begin{lstlisting}[style=leanappendix]
import Mathlib

theorem contractible_connected_and_trivial_homotopyGroups
    (X : Type*) [TopologicalSpace X] [ContractibleSpace X] :
    ConnectedSpace X ∧ ∀ (n : ℕ) (x : X), Subsingleton (HomotopyGroup (Fin n) X x) := by
  sorry
\end{lstlisting}

\centering
\footnotesize
\setlength{\tabcolsep}{6pt}
\begin{tabular}{lcccccc}
\toprule
\textbf{Compile} & \textbf{GPT} & \textbf{Gemini} & \textbf{R1} & \textbf{R2} & \textbf{R3} & \textbf{R4} \\
\midrule
1 & 9 & 5 & 10 & 10 & 9 & 10 \\
\bottomrule
\end{tabular}
\par\smallskip
\raggedright
\textbf{Audit note.} All available reviewers accepted the high-level Lean encoding via \texttt{ContractibleSpace}, \texttt{ConnectedSpace}, and subsingleton \texttt{HomotopyGroup}s.
\end{tcolorbox}

\begin{tcolorbox}[title={Case B-041: semantic-looking output blocked by compilation}, colback=white, colframe=gray, breakable]
\small
\textbf{Original NL statement.}
Let \(X\) and \(Y\) be topological spaces with basepoints \(x_0\in X\) and \(y_0\in Y\). Prove that \(\pi_1(X\times Y,(x_0,y_0))\cong \pi_1(X,x_0)\times \pi_1(Y,y_0)\).

\textbf{Translated Lean 4 code.}
\begin{lstlisting}[style=leanappendix]
import Mathlib

open CategoryTheory FundamentalGroupoid FundamentalGroupoidFunctor

noncomputable def «benmckay_top_fundamental-groups_336»
    (X : Type*) (Y : Type*) [TopologicalSpace X] [TopologicalSpace Y]
    (x₀ : X) (y₀ : Y) :
    FundamentalGroup (X × Y) (x₀, y₀) ≃* FundamentalGroup X x₀ × FundamentalGroup Y y₀ where
  toFun p := (Path.Homotopic.projLeft p, Path.Homotopic.projRight p)
  invFun q := Path.Homotopic.prod q.1 q.2
  left_inv p := Path.Homotopic.prod_projLeft_projRight p
  right_inv q := by
    ext
    · exact Path.Homotopic.projLeft_prod q.1 q.2
    · exact Path.Homotopic.projRight_prod q.1 q.2
  map_mul' a b := by
    set A := TopCat.of X
    set B := TopCat.of Y
    set pL := projLeft A B
    set pR := projRight A B
    show (pL.map (b ≫ a), pR.map (b ≫ a)) =
      (pL.map b ≫ pL.map a, pR.map b ≫ pR.map a)
    rw [pL.map_comp, pR.map_comp]
\end{lstlisting}

\centering
\footnotesize
\setlength{\tabcolsep}{6pt}
\begin{tabular}{lcccccc}
\toprule
\textbf{Compile} & \textbf{GPT} & \textbf{Gemini} & \textbf{R1} & \textbf{R2} & \textbf{R3} & \textbf{R4} \\
\midrule
0 & 2 & 3 & 10 & 0 & 0 & 10 \\
\bottomrule
\end{tabular}
\par\smallskip
\raggedright
\textbf{Audit note.} The declaration attempts the intended isomorphism but fails the compiler filter. Two reviewers followed the compile-aware rubric and assigned 0; R4 judged the mathematical encoding as matching (R1 scored the agent's output).
\end{tcolorbox}

\begin{tcolorbox}[title={Case B-091: consensus-positive boundary case}, colback=white, colframe=gray, breakable]
\small
\textbf{Original NL statement.}
Show that the map \(\varphi:G \to G \times H\) given by \(\varphi(x) = (x,1)\) is an injective homomorphism.

\textbf{Translated Lean 4 code.}
\begin{lstlisting}[style=leanappendix]
import Mathlib

theorem srdoty_alg_products_372
    (G H : Type*) [Group G] [Group H] :
    Function.Injective (MonoidHom.inl G H) := by sorry
\end{lstlisting}

\centering
\footnotesize
\setlength{\tabcolsep}{6pt}
\begin{tabular}{lcccccc}
\toprule
\textbf{Compile} & \textbf{GPT} & \textbf{Gemini} & \textbf{R1} & \textbf{R2} & \textbf{R3} & \textbf{R4} \\
\midrule
1 & 10 & 10 & 10 & 7 & 10 & 6 \\
\bottomrule
\end{tabular}
\par\smallskip
\raggedright
\textbf{Audit note.} Human majority rejected the consensus-positive label (scores 7, 10, 6): the statement states \texttt{Function.Injective (MonoidHom.inl G H)}, with the homomorphism property supplied by the type of Mathlib's bundled homomorphism $x\mapsto(x,1)$ rather than stated as part of the conclusion. The case turns on whether a property supplied by a type counts as stated; we keep the majority label, and counting the output as faithful would raise Batch-B agreement to 55/59.
\end{tcolorbox}

\begin{tcolorbox}[title={Precision B-022: ring sign identity}, colback=white, colframe=gray, breakable]
\small
\textbf{Original NL statement.}
Let \(R\) be a ring and let \(a\in R\). Show that \((-1)a = -a\).

\textbf{Translated Lean 4 code.}
\begin{lstlisting}[style=leanappendix]
import Mathlib

theorem srdoty_alg_Z_n_513 {R : Type*} [Ring R] (a : R) : (-1) * a = -a := by sorry
\end{lstlisting}

\centering
\footnotesize
\setlength{\tabcolsep}{6pt}
\begin{tabular}{lcccccc}
\toprule
\textbf{Compile} & \textbf{GPT} & \textbf{Gemini} & \textbf{R1} & \textbf{R2} & \textbf{R3} & \textbf{R4} \\
\midrule
1 & 10 & 10 & 10 & 10 & 10 & 10 \\
\bottomrule
\end{tabular}
\par\smallskip
\raggedright
\textbf{Audit note.} The declaration is a literal Lean rendering of the textbook ring identity.
\end{tcolorbox}

\begin{tcolorbox}[title={Precision B-032: linear congruence with a unit coefficient}, colback=white, colframe=gray, breakable]
\small
\textbf{Original NL statement.}
Let \(n>1\) be an integer. Prove that the linear congruence \(ax \equiv b \pmod{n}\) has a solution in \(\mathbb{Z}_n\) whenever the class \([a] \in \mathbb{Z}_n\) is a unit.

\textbf{Translated Lean 4 code.}
\begin{lstlisting}[style=leanappendix]
import Mathlib

theorem srdoty_alg_Z_n_1021 (n : ℕ) (hn : n > 1)
    (a b : ZMod n) (ha : IsUnit a) : ∃ x : ZMod n, a * x = b := by sorry
\end{lstlisting}

\centering
\footnotesize
\setlength{\tabcolsep}{6pt}
\begin{tabular}{lcccccc}
\toprule
\textbf{Compile} & \textbf{GPT} & \textbf{Gemini} & \textbf{R1} & \textbf{R2} & \textbf{R3} & \textbf{R4} \\
\midrule
1 & 10 & 10 & 10 & 10 & 10 & 10 \\
\bottomrule
\end{tabular}
\par\smallskip
\raggedright
\textbf{Audit note.} The Lean statement works directly in \texttt{ZMod n} and states solvability of \(ax=b\) for a unit \(a\), matching the source claim about \(\mathbb{Z}_n\).
\end{tcolorbox}

\begin{tcolorbox}[title={Precision B-050: rearranging an absolutely convergent series}, colback=white, colframe=gray, breakable]
\small
\textbf{Original NL statement.}
Suppose \(\sum_{n=1}^\infty a_n\) converges absolutely. Show that for any rearrangement of the series, the sum remains the same.

\textbf{Translated Lean 4 code.}
\begin{lstlisting}[style=leanappendix]
import Mathlib

theorem jirilebl_ra_ch_seq_ser_2828
    (a : ℕ → ℝ) (_ha : Summable fun n => |a n|) (σ : Equiv.Perm ℕ) :
    ∑' n, a (σ n) = ∑' n, a n := by sorry
\end{lstlisting}

\centering
\footnotesize
\setlength{\tabcolsep}{6pt}
\begin{tabular}{lcccccc}
\toprule
\textbf{Compile} & \textbf{GPT} & \textbf{Gemini} & \textbf{R1} & \textbf{R2} & \textbf{R3} & \textbf{R4} \\
\midrule
1 & 9 & 10 & 10 & 10 & 10 & 9 \\
\bottomrule
\end{tabular}
\par\smallskip
\raggedright
\textbf{Audit note.} Absolute convergence is encoded as summability of \(\lvert a_n\rvert\) and a rearrangement as a permutation of \(\mathbb{N}\). One reviewer noted the indexing from \(0\) rather than \(1\), which does not change the claim.
\end{tcolorbox}

\begin{tcolorbox}[title={Precision B-056: maps from compact to Hausdorff spaces are proper}, colback=white, colframe=gray, breakable]
\small
\textbf{Original NL statement.}
Prove that every continuous map \(f \colon X \to Y\) from a compact space \(X\) to a Hausdorff space \(Y\) is proper.

\textbf{Translated Lean 4 code.}
\begin{lstlisting}[style=leanappendix]
import Mathlib

theorem benmckay_top_continuity_310
    {X Y : Type*} [TopologicalSpace X] [TopologicalSpace Y]
    [CompactSpace X] [T2Space Y]
    {f : X → Y} (hf : Continuous f) : IsProperMap f := by sorry
\end{lstlisting}

\centering
\footnotesize
\setlength{\tabcolsep}{6pt}
\begin{tabular}{lcccccc}
\toprule
\textbf{Compile} & \textbf{GPT} & \textbf{Gemini} & \textbf{R1} & \textbf{R2} & \textbf{R3} & \textbf{R4} \\
\midrule
1 & 10 & 10 & -- & 10 & 10 & 10 \\
\bottomrule
\end{tabular}
\par\smallskip
\raggedright
\textbf{Audit note.} The Lean statement uses Mathlib's \texttt{IsProperMap} under \texttt{CompactSpace} and \texttt{T2Space} hypotheses, a direct rendering of the source claim.
\end{tcolorbox}

\begin{tcolorbox}[title={Precision B-094: Schwarz lemma bound}, colback=white, colframe=gray, breakable]
\small
\textbf{Original NL statement.}
Suppose \(f \colon \mathbb{D} \to \mathbb{D}\) is holomorphic and \(f(0) = 0\), then \(\lvert f(z)\rvert \leq \lvert z\rvert\) for all \(z \in \mathbb{D}\).

\textbf{Translated Lean 4 code.}
\begin{lstlisting}[style=leanappendix]
import Mathlib

open Complex Metric Set

theorem jirilebl_ca_ca_5555
    (f : ℂ → ℂ)
    (hf_diff : DifferentiableOn ℂ f (ball 0 1))
    (hf_maps : MapsTo f (ball 0 1) (ball 0 1))
    (hf_zero : f 0 = 0)
    (z : ℂ)
    (hz : z ∈ ball (0 : ℂ) 1) :
    ‖f z‖ ≤ ‖z‖ := by sorry
\end{lstlisting}

\centering
\footnotesize
\setlength{\tabcolsep}{6pt}
\begin{tabular}{lcccccc}
\toprule
\textbf{Compile} & \textbf{GPT} & \textbf{Gemini} & \textbf{R1} & \textbf{R2} & \textbf{R3} & \textbf{R4} \\
\midrule
1 & 10 & 10 & -- & 10 & 10 & 9 \\
\bottomrule
\end{tabular}
\par\smallskip
\raggedright
\textbf{Audit note.} The unit disk is encoded as \texttt{ball 0 1}, holomorphy as complex differentiability on it, and the self-map condition with \texttt{MapsTo}; the conclusion is the stated bound.
\end{tcolorbox}

\subsection{Four-Reviewer Theorem-Name Overlap Diagnostic}
\label{app:four_reviewer_overlap}

Some theorems appear both in the Reviewer 1 audit and in Batches A and B. Table~\ref{tab:four_reviewer_overlap} lists the 20 such theorems on which Reviewer 1 and all three other reviewers gave numeric scores (3 in Batch A, 17 in Batch B). For two Batch-A rows (A-007, A-050), all four reviewers scored the same agent declaration, shown to Batch A without its comments; A-049 is a different statement that shares its theorem name. For the 17 Batch-B rows, Reviewer 1 scored our agent's output, while the other three reviewers scored Aristotle's output for a statement with the same theorem name. This diagnostic records which outputs each reviewer scored for the same theorem; it is not an inter-annotator agreement, since most rows compare scores of different outputs.

\begin{table}[H]
\centering
\caption{\textbf{Theorem-name overlaps with numeric scores from Reviewers 1--4.}
Scores $\geq 9$ are faithful. \textbf{Maj4}=1 means at least three of the four reviewers scored the theorem-name occurrence as faithful; \textbf{All4}=1 means all four did. R1 scored our agent's output; R2--R4 scored the Batch-A or Batch-B output.}
\label{tab:four_reviewer_overlap}
\scriptsize
\setlength{\tabcolsep}{2.5pt}
\begin{tabularx}{\linewidth}{ll>{\raggedright\arraybackslash}Xrrrrcc}
\toprule
\textbf{Audit} & \textbf{Case} & \textbf{Theorem name} & \textbf{R1} & \textbf{R2} & \textbf{R3} & \textbf{R4} & \textbf{Maj4} & \textbf{All4} \\
\midrule
A & case\_007 & \texttt{jirilebl\_\allowbreak{}ca\_\allowbreak{}ca\_\allowbreak{}1436} & 10 & 10 & 10 & 9 & 1 & 1 \\
A & case\_049 & \texttt{jirilebl\_\allowbreak{}ra\_\allowbreak{}ch-seq-funcs\_\allowbreak{}497} & 9 & 2 & 4 & 5 & 0 & 0 \\
A & case\_050 & \texttt{benmckay\_\allowbreak{}top\_\allowbreak{}covering-spaces\_\allowbreak{}1191} & 10 & 10 & 9 & 10 & 1 & 1 \\
B & case\_011 & \texttt{jirilebl\_\allowbreak{}ra\_\allowbreak{}ch-multivar-int\_\allowbreak{}2082} & 10 & 10 & 10 & 9 & 1 & 1 \\
B & case\_012 & \texttt{jirilebl\_\allowbreak{}ca\_\allowbreak{}ca\_\allowbreak{}9443} & 10 & 7 & 10 & 10 & 1 & 0 \\
B & case\_022 & \texttt{srdoty\_\allowbreak{}alg\_\allowbreak{}Z\_\allowbreak{}n\_\allowbreak{}513} & 10 & 10 & 10 & 10 & 1 & 1 \\
B & case\_025 & \texttt{jirilebl\_\allowbreak{}ra\_\allowbreak{}ch-seq-ser\_\allowbreak{}4420} & 9 & 10 & 10 & 9 & 1 & 1 \\
B & case\_032 & \texttt{srdoty\_\allowbreak{}alg\_\allowbreak{}Z\_\allowbreak{}n\_\allowbreak{}1021} & 10 & 10 & 10 & 10 & 1 & 1 \\
B & case\_041 & \texttt{benmckay\_\allowbreak{}top\_\allowbreak{}fundamental-groups\_\allowbreak{}336} & 10 & 0 & 0 & 10 & 0 & 0 \\
B & case\_042 & \texttt{srdoty\_\allowbreak{}alg\_\allowbreak{}Z\_\allowbreak{}n\_\allowbreak{}505} & 10 & 10 & 10 & 10 & 1 & 1 \\
B & case\_045 & \texttt{srdoty\_\allowbreak{}alg\_\allowbreak{}actions\_\allowbreak{}705} & 10 & 10 & 10 & 10 & 1 & 1 \\
B & case\_049 & \texttt{srdoty\_\allowbreak{}alg\_\allowbreak{}abstract-gps\_\allowbreak{}942} & 10 & 10 & 10 & 10 & 1 & 1 \\
B & case\_050 & \texttt{jirilebl\_\allowbreak{}ra\_\allowbreak{}ch-seq-ser\_\allowbreak{}2828} & 10 & 10 & 10 & 9 & 1 & 1 \\
B & case\_069 & \texttt{jirilebl\_\allowbreak{}ca\_\allowbreak{}ca\_\allowbreak{}2441} & 10 & 10 & 10 & 10 & 1 & 1 \\
B & case\_074 & \texttt{benmckay\_\allowbreak{}top\_\allowbreak{}fundamental-groups\_\allowbreak{}24} & 10 & 10 & 10 & 10 & 1 & 1 \\
B & case\_081 & \texttt{jirilebl\_\allowbreak{}ra\_\allowbreak{}ch-seq-ser\_\allowbreak{}4256} & 10 & 5 & 10 & 10 & 1 & 0 \\
B & case\_085 & \texttt{jirilebl\_\allowbreak{}ra\_\allowbreak{}ch-seq-funcs\_\allowbreak{}497} & 9 & 10 & 10 & 10 & 1 & 1 \\
B & case\_090 & \texttt{srdoty\_\allowbreak{}alg\_\allowbreak{}Z\_\allowbreak{}n\_\allowbreak{}725} & 10 & 9 & 10 & 10 & 1 & 1 \\
B & case\_091 & \texttt{srdoty\_\allowbreak{}alg\_\allowbreak{}products\_\allowbreak{}372} & 10 & 7 & 10 & 6 & 0 & 0 \\
B & case\_100 & \texttt{srdoty\_\allowbreak{}alg\_\allowbreak{}Z\_\allowbreak{}n\_\allowbreak{}828} & 10 & 10 & 10 & 10 & 1 & 1 \\
\bottomrule
\end{tabularx}
\end{table}

\section{Additional Factorial and Tool-Usage Details}
\label{app:factorial_extra}

\begin{table}[H]
\centering
\caption{\textbf{Validity vs.\ semantic selectivity by tool configuration.}
Conditional faithfulness is Faithful/Compile. The gap column counts outputs that compile but fail strict semantic consensus.}
\label{tab:factorial_boundary}
{\scriptsize
\setlength{\tabcolsep}{5pt}
\renewcommand{\arraystretch}{1.05}
\begin{tabular}{@{}ccc|cccc@{}}
\toprule
\multicolumn{3}{c|}{\textbf{Config}} & \textbf{Compile} & \textbf{Faithful} & \textbf{Faithful|Compile} & \textbf{Compile-pass gap} \\
\textbf{T} & \textbf{F} & \textbf{S} & \textbf{Count} & \textbf{Count} & \textbf{Rate} & \textbf{Count} \\
\midrule
0 & 0 & 0 & 54 & 45  & 83.3\% & 9 \\
1 & 0 & 0 & 65 & 56  & 86.2\% & 9 \\
0 & 0 & 1 & 109 & 80 & 73.4\% & 29 \\
1 & 0 & 1 & 121 & 84 & 69.4\% & 37 \\
\midrule
0 & 1 & 0 & 203 & 127 & 62.6\% & 76 \\
0 & 1 & 1 & 189 & 132 & 69.8\% & 57 \\
1 & 1 & 0 & 207 & 122 & 58.9\% & 85 \\
1 & 1 & 1 & 198 & 131 & 66.2\% & 67 \\
\bottomrule
\end{tabular}
}
\end{table}

\begin{table}[H]
\centering
\caption{\textbf{Tool usage counts.}
Cumulative tool invocations over the 227 statements.}
\label{tab:tool_usage}
\setlength{\tabcolsep}{5pt}
\renewcommand{\arraystretch}{1.1}
{\scriptsize
\begin{tabular}{@{}c|rrr|l@{}}
\toprule
\textbf{$(T,F,S)$} & \textbf{Trans.} & \textbf{REPL} & \textbf{$S_{\text{total}}$} & \textbf{Notes} \\
\midrule
(0,1,0) & 0   & 899 & 0    & Feedback-only \\
(0,1,1) & 0   & 641 & 1157 & +Search (REPL$\downarrow$) \\
\midrule
(1,1,0) & 148 & 864 & 0    & Feedback + Expert \\
(1,1,1) & 48 & 586 & 1274 & +Search (REPL$\downarrow$) \\
\bottomrule
\end{tabular}
}
\end{table}

\begin{table}[H]
\centering
\caption{\textbf{Trajectory length and faithfulness.}
Rows pool runs by final trajectory length. $F{=}1$ pools configs 010, 011, 110, and 111; $F{=}0$ pools the no-feedback configs with tools, 001, 100, and 101. Accuracy follows the paper's per-entry consensus metric.}
\label{tab:trajectory_cliff}
\footnotesize
\setlength{\tabcolsep}{7pt}
\begin{tabular}{lcccc}
\toprule
\textbf{Step bin} & \multicolumn{2}{c}{$\boldsymbol{F{=}1}$} & \multicolumn{2}{c}{$\boldsymbol{F{=}0}$, nonempty} \\
& \textbf{Faithful} & \textbf{Rate} & \textbf{Faithful} & \textbf{Rate} \\
\midrule
1--2   & 165/199 & 82.9\% & 67/194 & 34.5\% \\
3--5   & 136/199 & 68.3\% & 72/187 & 38.5\% \\
6--9   & 113/180 & 62.8\% & 32/110  & 29.1\% \\
10--12 & 46/92  & 50.0\% & 14/43   & 32.6\% \\
13--18 & 34/81  & 42.0\% & 21/73  & 28.8\% \\
19--24 & 18/157  & 11.5\% & 14/74  & 18.9\% \\
\bottomrule
\end{tabular}
\end{table}

\section{\texorpdfstring{Validating the Review of Rejected Outputs}{Validating the Review of Rejected Outputs}}
\label{app:review_validation}
\paragraph{Outputs.} The validation set has every compiling output that GPT-5.2 rejects (grade below 9, the rule of Section~\ref{sec:newer} and Figure~\ref{fig:judge_regime}) and that has a human-majority label: 64 outputs of our agent (Batch A) and 5 of Aristotle (Batch B), 69 in all, on 66 statements; the human majority judges 57 of them unfaithful. The one-shot set has every compiling output of the seven one-shot systems that GPT-5.2 rejects, on statements whose recorded input matches the benchmark: 142 outputs on 105 statements. For Table~\ref{tab:gap_real}, which uses the criterion of Figure~\ref{fig:main_results}, we also reviewed the three compiling outputs that only Gemini-2.5-Pro rejects (Goedel-Prover on BC007, one-shot GPT-5.2 on BC129 and BC130); their first stage also saw Gemini's grade and reason, and all three are translation errors.

The validation results are in Tables~\ref{tab:review_validation} and~\ref{tab:review_parity} (Section~\ref{sec:review}).

\paragraph{Procedure.} Both sets use the instructions of our first review run, on the later models (Section~\ref{sec:newer}; Appendix~\ref{app:newer}), with three changes. The first stage names the systems under review. It may read only its input file, the textbook sources, and its own Lean checks, so that the human labels, which are stored elsewhere on the same machine, stay hidden; in the validation run, the agents' logs show that no other file was opened. It sees only rejected outputs, so there is no accepted output of another model to compare with. The stages run in order: Claude Opus 5.5 agents in Claude Code with the Lean REPL review batches of 12 statements; GPT-6.1 Sol reviews each output with the first review in view; and GPT-6.1 Sol decides the outputs on which exactly one review finds a translation error. Both stages see the statement, its source excerpt, the output, and the grades and stated reasons of GPT-5.2 and of Claude Sonnet~5, as for the later models; for Aristotle and the one-shot systems we obtained the Sonnet~5 grades with the paper's judge prompt. For the three outputs that only Gemini-2.5-Pro rejects, the first stage also sees Gemini's grade and reason. No stage sees a human label. Final classes follow the rule of the later-model review, which reproduces its released classes exactly. The prompts, both reviews, the discussion, and the reviewers' Lean checks are released.

\paragraph{Details.} Over all 69 outputs, the review agrees with the human majority on 79.7\% ($\kappa=0.41$); pairs of reviewers agree on 69.6--74.6\% ($\kappa=0.12$--$0.27$). Of the four outputs that the review flags and the majority accepts, two are Lean-verified errors: A-008 (BC148) omits $a\le b$, and its statement is false for $b<a$; A-094 (BC019) asks for an isomorphism with the multiplicative monoid of \texttt{ZMod (2p)}, which no group has, so it is false for cyclic groups. A-027 (BC160) replaces Riemann with Lebesgue integrability, and A-012 (BC072) admits an extra case with \texttt{Nat.card} $=0$. Of the ten majority-unfaithful outputs that the review does not classify as translation errors, the first stage had flagged seven as translation errors and the discussion round overruled it, which is why T by either stage finds 54 of the 57; two are on statements that the release flags as defective (BC142, BC208), and two received grades of 8 from two of the three reviewers (BC111, BC156).

\begin{table}[H]
\centering
\footnotesize
\setlength{\tabcolsep}{4pt}
\caption{\textbf{The review of the seven one-shot systems' rejections} (GPT-5.2 alone; the three criterion-only rejections, all T, are not included). $R$: rejected compiling outputs; T: translation errors (final class, and by either stage); J: judge errors; B: defective statements; $e\geq$: translation errors over compiling outputs, a lower bound on the unfaithful share.}
\label{tab:earlier_review}
\begin{tabular}{lrrrrrr}
\toprule
\textbf{System} & $R$ & T & T (either) & J & B & $e\geq$ \\
\midrule
GPT-5.2 & 7 & 5 & 6 & 1 & 1 & 9.3\% \\
Gemini-2.5-Pro & 3 & 1 & 3 & 0 & 2 & 1.5\% \\
Herald & 26 & 23 & 24 & 2 & 1 & 46.9\% \\
Kimina-AutoF & 46 & 44 & 45 & 1 & 1 & 60.3\% \\
StepFun & 14 & 11 & 12 & 2 & 1 & 25.0\% \\
Kimina-Prover & 14 & 14 & 14 & 0 & 0 & 58.3\% \\
Goedel-Prover & 32 & 31 & 31 & 1 & 0 & 66.0\% \\
\midrule
Total & 142 & 129 & 135 & 7 & 6 & \\
\bottomrule
\end{tabular}
\end{table}

\section{\texorpdfstring{The Later Models' Rejections}{The Later Models' Rejections}}
\label{app:newer}

This appendix details the review of the 54 compiling outputs of GPT-6 Astra and Claude Opus 5.5 that GPT-5.2 rejects (Section~\ref{sec:newer}). This was our first review run. It follows the procedure of Section~\ref{sec:review}, except that the first stage also saw the other model's output for the same statement when GPT-5.2 accepted it (Appendix~\ref{app:review_validation} lists the differences); the discussion round decided the six outputs on which the two stages disagreed about a translation error.

\subsection{What the Review Finds}
The review classifies 32 of the 50 rejections that are not translation errors as judge errors, 11 as defective benchmark items, and 7 as partly both (Table~\ref{tab:judge_failures}). By its account, GPT-5.2 rejects equivalent but non-standard encodings, such as $\mathrm{O}(n)$ with membership in $\mathrm{GL}(n)$ written as \texttt{IsUnit}. It grades nearly identical outputs differently: in 10 cases it accepts an essentially identical statement from the other model, with grades as far apart as 10 and 2. Its stated reason is sometimes wrong; for example, it claims that a product of two transpositions is the inverse of the 3-cycle it actually equals. It also penalizes literal translations of statements that are false as written, and the minimal repairs that make them true. Table~\ref{tab:judge_examples} shows one example each of an identical output graded differently, a wrong reason, and a rejected encoding.

\begin{table}[H]
\centering
\caption{\textbf{Model-based review of the later models' rejections:} main failure of the judge for the 50 rejected compiling outputs that the review does not classify as translation errors.}
\label{tab:judge_failures}
\footnotesize
\begin{tabularx}{\linewidth}{@{}Xr@{}}
\toprule
\textbf{Failure of the judge} & \textbf{Outputs} \\
\midrule
Rejects an equivalent but non-standard encoding & 20 \\
Penalizes a literal translation, or a repair, of a defective statement & 11 \\
Accepts an essentially identical statement from the other model & 10 \\
Gives a mathematically wrong reason & 9 \\
\bottomrule
\end{tabularx}
\end{table}

\begin{table}[H]
\centering
\caption{\textbf{Three rejected later-model outputs that the review classifies as faithful.} Statements and code are abridged; grades are out of 10. The review records are in the release.}
\label{tab:judge_examples}
\footnotesize
\renewcommand{\arraystretch}{1.15}
\begin{tabularx}{\linewidth}{@{}>{\raggedright\arraybackslash}p{0.2\linewidth}>{\raggedright\arraybackslash}p{0.25\linewidth}>{\raggedright\arraybackslash}p{0.22\linewidth}>{\raggedright\arraybackslash}X@{}}
\toprule
\textbf{Statement (abridged)} & \textbf{Lean output (abridged)} & \textbf{GPT-5.2's objection} & \textbf{Why the rejection is wrong} \\
\midrule
\textbf{BC181.} If $X$ is Hausdorff and continuous $f,g\colon X\to Y$ agree on a dense subset, then $f=g$. &
\texttt{[T2Space X]}, \texttt{Dense s}, \texttt{Set.EqOn f g s} $\vdash$ \texttt{f = g} (GPT-6 Astra; Opus 5.5 writes the same). &
Grade 4: the result needs $Y$, not $X$, to be Hausdorff. &
The objection concerns the statement, which copies a slip in its source; the output translates it literally, and Opus's essentially identical output receives 10. \\
\midrule
\textbf{BC062.} With $\alpha=(1,2)$ and $\beta=(1,2,3)$, $\langle\alpha,\beta\rangle=S_3$. &
$\beta$ \texttt{:= swap 0 1 * swap 1 2} and \texttt{closure} $\{\alpha,\beta\}=\top$ (GPT-6 Astra). &
Grade 8: in Mathlib this product is the inverse of the 3-cycle. &
In Mathlib $(f*g)(x)=f(g(x))$, so $\beta$ maps $0\mapsto1\mapsto2\mapsto0$, which is $(1,2,3)$ after relabeling; \texttt{decide} confirms it. \\
\midrule
\textbf{BC042.} $\mathrm{O}(n)=\{A\in\mathrm{GL}(n): AA^{\mathsf T}=I\}$. &
$\{A \mid A^{\mathsf T}A=1\}=\{A \mid \texttt{IsUnit}\,A\wedge AA^{\mathsf T}=1\}$ (GPT-6 Astra). &
Grade 6: it does not present $\mathrm{O}(n)$ or $\mathrm{GL}(n)$ as groups, and uses $A^{\mathsf T}A$ on the left. &
$A^{\mathsf T}A=1$ characterizes $\mathrm{O}(n)$, since it forces $A^{-1}=A^{\mathsf T}$, and \texttt{IsUnit} is membership in $\mathrm{GL}(n)$; the one-sided identity is what the statement asks to prove. \\
\bottomrule
\end{tabularx}
\end{table}

\subsection{Rejection Shares and the Judge's Rejection Rate}
Real errors make up 83\% of GPT-5.2's rejections for our agent (53 of 64) and 80\% for Aristotle (4 of 5), by human audit, and 71--100\% for six of the seven one-shot systems (1 of 3 for Gemini-2.5-Pro), by our review, but 10\% for Opus 5.5 (3 of 29) and 4\% for GPT-6 Astra (1 of 25), by our review (14\% and 16\% counting every output that either stage flags). The rate at which the judge rejects outputs that are not translation errors does not fall with them: it is an estimated 7.9\% for our agent and 1 of 47 (2.1\%) for Aristotle, and, by the review's labels, at least 11.3\% for GPT-6 Astra and 12.1\% for Opus 5.5 (7.8\% for each, counting only judge errors, J). By the human audits, an estimated 30\% of our agent's compiling outputs and 11\% of Aristotle's are unfaithful; the errors found among the later models' rejections are 0.5\% and 1.4\% of their compiling outputs, a lower bound on their unfaithful share.

\subsection{A Second Judge and Re-Grading}
\label{app:newer_regrade}
Requiring Sonnet~5 to reject as well keeps all four translation errors and 29 of the other 49 graded rejections, so the share of real errors rises from 7\% to 12\%. With five more GPT-5.2 calls per rejected output, 37 of the 50 rejections that are not translation errors stay rejected, a majority accepts 2 of the 4 errors, and 4 of 40 sampled accepted outputs flip to rejection.

\subsection{Checks in Lean}
\label{app:lean_evidence}
Part of this can be checked without trusting any reviewer. Lean-checked certificates cover 15 of the rejected outputs; the three kinds of evidence overlap and show different things. For 9 of the 10 pairs in which GPT-5.2 accepts one model's output and rejects the other's (Table~\ref{tab:judge_failures}), Lean proves the two statements equivalent, by definitional unfolding or a direct correspondence of their hypotheses and conclusions, yet GPT-5.2 grades them differently; one pair is textually identical and graded 10 and 2. For 4 of the 9 rejections with a wrong stated reason, Lean refutes that reason, which does not by itself show the rest of the output faithful. For 4 rejected outputs, Lean proves equivalence to an output that human reviewers accepted for the same statement. All 15 are among the outputs that the review classifies as not translation errors.

Claude Opus 5.5 agents wrote the Lean proofs; a fixed checker decides whether they count. A checked equivalence shows that two statements have the same truth conditions; it supports, but does not by itself establish, that they make the same claim as the natural-language statement. The checker places each output verbatim, without its import line, in its own namespace (\texttt{R} for the rejected output, \texttt{P} for the other model's accepted output or for the human-accepted output), so the agents cannot change the statements. For an equivalence, the certificate must have type \texttt{R}-statement $\leftrightarrow$ \texttt{P}-statement up to universe instantiation and definitional unfolding, both directions must be functions that use their hypothesis, and \texttt{\#print axioms} must show no \texttt{sorryAx} and nothing beyond \texttt{propext}, \texttt{Classical.choice}, and \texttt{Quot.sound}. For a stated reason, every recorded claim must compile under the same axiom check. Table~\ref{tab:lean_evidence} summarizes the outcomes; the release contains every proof and check record.

\begin{table}[H]
\centering
\caption{\textbf{Lean checks of the 54 rejected compiling outputs of the later models.}}
\label{tab:lean_evidence}
\footnotesize
\setlength{\tabcolsep}{3pt}
\begin{tabularx}{\linewidth}{@{}>{\raggedright\arraybackslash}Xrr@{}}
\toprule
\textbf{Check} & \textbf{Items} & \textbf{Verified} \\
\midrule
Equivalent to the other model's accepted output & 10 & 9 \\
Judge's stated reason refuted & 9 & 4 \\
Equivalent to a human-accepted output & 9 & 4 \\
\midrule
Distinct rejections covered & & 15 \\
\bottomrule
\end{tabularx}
\end{table}

\paragraph{Outcomes that are not verifications.} In the first group, the pair for BC205 differs in its hypotheses (one output requires covering maps to be surjective) and is not proved equivalent. In the second group, three reasons concern whether a definition matches the textbook's and are not Lean facts (BC135 and both outputs for BC209), and two would need Lebesgue's criterion for Riemann integrability, which Mathlib does not contain (BC155, BC158); these five reasons were not decided by this check. In the third group, equivalence was not established for four pairs: Astra's BC092 output reads the angle as unoriented, which makes the statement false outside $[0,\pi]$, as in its source, while the human-accepted output uses the oriented angle; the human-accepted output for BC042 generalizes the statement to every commutative ring; the human-accepted output for BC160 uses Lebesgue integrability (\texttt{IntervalIntegrable}) where the statement concerns the Riemann integral; and the human-accepted output for BC180 translates literally a statement that is false as written, while the rejected output adds the nonemptiness that makes it true. BC046 could not be run through the checker.

\subsection{Supplying Definitions to the Judge}
\label{app:definitions}
Giving the judge a fixed block of library definitions does not change its rejections of outputs that humans judge faithful. GPT-5.2 received the Mathlib definitions that each output uses, extracted from the Lean environment by a fixed rule and capped in length, with the prompt otherwise unchanged. On the 74 human-labeled rejections of our agent and Aristotle, with three calls per output and condition, the judge accepts by majority the same 8 of the 16 human-faithful outputs with and without the block, and 1 and 0 of the 58 human-unfaithful ones. On the later models, acceptance of the 32 outputs that the review classes as judge errors goes from 10 to 11, with five outputs changing in either direction.

For each output, a metaprogram elaborates the code on top of Mathlib and collects the constants used by the types of its declarations and by the values of its definitions. We keep constants from Mathlib modules only (core constants such as \texttt{Eq}, \texttt{Nat}, or \texttt{HMul.hMul} are dropped), in order of first use and at most 30, and give each its signature, its docstring, and, for a definition, its value when short; the block is capped at 6000 characters, a limit reached for 41 of the 128 outputs. The judge receives the paper's system prompt and user message unchanged, with this block added after the Lean code. Each output is graded three times under each condition and counts as accepted if at least two grades are 9 or more; for the later models, the condition without definitions uses the first three re-grading calls of Appendix~\ref{app:newer_regrade}.

\begin{table}[H]
\centering
\caption{\textbf{Outputs accepted by majority without (A) and with (B) the Mathlib definitions.} Human labels for our agent and Aristotle; final review classes for the later models.}
\label{tab:definitions}
\footnotesize
\begin{tabular}{lrrr}
\toprule
\textbf{Rejected outputs} & \textbf{n} & \textbf{A} & \textbf{B} \\
\midrule
Agent and Aristotle, human-faithful & 16 & 8 & 8 \\
Agent and Aristotle, human-unfaithful & 58 & 1 & 0 \\
\midrule
Later, judge error (J) & 32 & 10 & 11 \\
Later, J/B & 7 & 3 & 2 \\
Later, defective item (B) & 11 & 1 & 1 \\
Later, translation error (T) & 4 & 1 & 0 \\
\bottomrule
\end{tabular}
\end{table}

Item by item, no human-faithful output changes; one human-unfaithful output is accepted only without the block; among the later models' judge-error class, two outputs are accepted only without the block and three only with it. The mean grade changes from 7.92 to 7.77 on the human-faithful outputs and from 4.86 to 4.67 on the human-unfaithful ones.

\subsection{Defective Statements}
Of the 11 rejections on defective statements (Table~\ref{tab:judge_failures}), four concern statements that are false as written, as in their sources, and here the two later models differ. Opus adds the hypotheses that make three of them true: nonempty factors for the compactness of $X\times Y$ (\hyperref[tab:flagged_list]{BC179}, \hyperref[tab:flagged_list]{BC180}) and a Hausdorff $X$ for the one-point compactification (\hyperref[tab:flagged_list]{BC218}). GPT-5.2 rejects these repairs and gives GPT-6 Astra's literal translations 10. On the fourth (\hyperref[tab:flagged_list]{BC092}; Table~\ref{tab:flagged_list} lists all flagged statements), it does the opposite, rejecting Astra's literal statement and accepting Opus's repair. Our review counts both the literal translation and the minimal repair as not translation errors; GPT-5.2 follows neither policy consistently. Without the 18 flagged statements, GPT-5.2 reports gaps of 7.2 and 9.6 points for GPT-6 Astra and Opus 5.5, and the review classifies 1 of 15 and 3 of 20 of their rejections as translation errors; for our agent, humans confirm 45 of its 55 rejections (Appendix~\ref{app:flagged}).

\section{\texorpdfstring{Results without Flagged Statements}{Results without Flagged Statements}}
\label{app:flagged}
Table~\ref{tab:flagged} repeats the main measurements without the 18 statements listed in the release as false as written, extracted without needed context, or using an undefined term. Kimina Prover, Kimina AutoF, and StepFun are evaluated on the 200 of the remaining statements whose recorded input matches the benchmark statement.

Table~\ref{tab:flagged_list} lists the 18 statements and what is wrong with each.

\begin{table}[H]
\centering
\caption{\textbf{The 18 flagged statements.}}
\label{tab:flagged_list}
\footnotesize
\setlength{\tabcolsep}{3pt}
\begin{tabularx}{\linewidth}{@{}l>{\raggedright\arraybackslash}X@{}}
\toprule
\textbf{ID} & \textbf{Issue} \\
\midrule
\multicolumn{2}{@{}l}{\textit{False as written, as in the source}} \\
BC021 & ``Any proper normal subgroup'' must be nontrivial; the trivial subgroup is a counterexample. \\
BC092 & For the unoriented angle, the angle between $X$ and $\rho_\theta(X)$ equals $\theta$ only for $\theta\in[0,\pi]$. \\
BC179, BC180 & False when one factor is empty and the other is not compact. \\
BC181 & The Hausdorff hypothesis is needed on $Y$, not on $X$. \\
BC205 & Needs local path-connectedness; the Warsaw circle is a counterexample. \\
BC217 & Van Kampen needs path-connected triple intersections; the theta graph is a counterexample. \\
BC218 & Not a topology for non-Hausdorff $X$ unless the compact sets are also closed. \\
BC226 & Continuity is equivalent to $\overline{f^{-1}A}\subseteq f^{-1}\overline{A}$, not to equality. \\
\midrule
\multicolumn{2}{@{}l}{\textit{Context lost in extraction}} \\
BC130 & Restates the definition of conditional convergence, so it is a tautology. \\
BC142 & Refers to an interval $I$ ``as above'' that the statement never defines. \\
BC183 & The map to the fiber and the notation $X_{y_0}$ are not defined. \\
BC199 & Keeps only the trivial definedness half of the source exercise. \\
BC208 & Assumes the concatenations are defined, which makes the definedness claim trivial. \\
BC215 & The fiber $F$, the base point $y_0$, and the connecting maps are not defined. \\
BC216 & Drops the hypothesis that every $X_a$ contains $x_0$. \\
\midrule
\multicolumn{2}{@{}l}{\textit{Undefined term}} \\
BC044, BC048 & ``Matrix group'' is not defined: a subgroup or a closed subgroup of $\mathrm{GL}(n)$. \\
\bottomrule
\end{tabularx}
\end{table}

\begin{table}[H]
\centering
\caption{\textbf{Measured gap (points) on all statements and without the 18 flagged ones.} Rej.: rejected compiling outputs without the flagged statements; Err.: those that are translation errors. Upper rows: compile + GPT$\cap$Gemini criterion. Lower rows: acceptance by GPT-5.2 alone (Section~\ref{sec:newer}); errors by the human majority in Batch~A for the agent and after the discussion round for the later models.}
\label{tab:flagged}
\footnotesize
\setlength{\tabcolsep}{3pt}
\begin{tabular}{lrrrr}
\toprule
\textbf{System} & \textbf{All} & \textbf{Unflagged} & \textbf{Rej.} & \textbf{Err.} \\
\midrule
Kimina Prover & 6.5 & 7.0 & -- & -- \\
Goedel Prover & 14.5 & 14.4 & -- & -- \\
Kimina AutoF & 21.2 & 21.5 & -- & -- \\
Herald & 11.5 & 11.5 & -- & -- \\
StepFun & 6.5 & 5.5 & -- & -- \\
GPT-5.2 & 4.0 & 3.3 & -- & -- \\
Gemini-2.5-Pro & 1.3 & 1.4 & -- & -- \\
GPT-5.2 agent & 29.5 & 27.8 & -- & -- \\
\midrule
GPT-5.2 agent & 28.2 & 26.3 & 55 & 45 \\
GPT-6 Astra & 11.0 & 7.2 & 15 & 1 \\
Opus 5.5 & 12.8 & 9.6 & 20 & 3 \\
\bottomrule
\end{tabular}
\end{table}

\clearpage
\section{Additional Domain Results}
\label{app:domain_extra}

In this appendix, as in its tables, \emph{faithful} means accepted by the criterion (compiles, and both judges grade it at least 9) unless a human label is stated.

\subsection{\texorpdfstring{Domain Metrics of Our Agent (Configuration 111)}{Domain Metrics of Our Agent (Configuration 111)}}
\label{app:domain_111}

\begin{tcolorbox}[title=Domain metrics under config 111, colback=white, colframe=gray, breakable]
\small
\begin{table}[H]
\centering
\caption{\textbf{Domain metrics under config=111.}
Conditional faithfulness is computed as Faithful/Compile.}
\label{tab:domain_111_metrics}
\setlength{\tabcolsep}{4.5pt}
\resizebox{\linewidth}{!}{%
\begin{tabular}{lrrrrr}
\toprule
\textbf{Domain} & \textbf{Compile} & \textbf{Faithful} & \textbf{Faithful|Compile} & \textbf{Steps (mean)} & \textbf{Steps (median)} \\
\midrule
Complex Analysis (19) & 0.89 & 0.68 & 0.76 & 8.74 & 7 \\
Real Analysis (48)    & 0.92 & 0.56 & 0.61 & 9.98 & 7 \\
Algebra (97)          & 0.87 & 0.57 & 0.65 & 9.85 & 7 \\
Topology (63)         & 0.84 & 0.57 & 0.68 & 10.17 & 7 \\
\bottomrule
\end{tabular}
}
\end{table}
\end{tcolorbox}

\subsection{Judge Scores by Domain}
\label{app:domain_scores}

\begin{tcolorbox}[title=Average judge scores by domain across configurations, colback=white, colframe=gray, breakable]
\small
\begin{table}[H]
\centering
\caption{\textbf{Average GPT-5.2 faithfulness grade by domain and configuration.}
Columns correspond to tool configurations in $(T,F,S)$ bit order.}
\label{tab:domain_score_all}
\setlength{\tabcolsep}{4.5pt}
\begin{tabular}{lrrrrrrr}
\toprule
\textbf{Domain} & \textbf{001} & \textbf{010} & \textbf{011} & \textbf{100} & \textbf{101} & \textbf{110} & \textbf{111} \\
\midrule
Algebra          & 5.65 & 7.19 & 7.24 & 4.46 & 5.90 & 7.29 & 7.32 \\
Complex Analysis & 5.05 & 9.53 & 8.58 & 4.11 & 5.63 & 8.79 & 8.47 \\
Real Analysis    & 4.46 & 7.17 & 7.35 & 3.31 & 4.58 & 7.29 & 7.79 \\
Topology         & 5.29 & 6.84 & 7.03 & 4.29 & 5.83 & 6.87 & 7.11 \\
\bottomrule
\end{tabular}
\end{table}

\noindent\textbf{Interpretation.}
Complex Analysis attains the highest scores under feedback-enabled settings, consistent with its highest Faithful rate under config 111 (Appendix~\ref{app:domain_111}); Topology has the lowest scores in these settings.
\end{tcolorbox}

\subsection{Domain-Level Feedback Effects}
\label{app:domain_feedback}

\begin{figure}[H]
\centering
\includegraphics[width=0.7\textwidth]{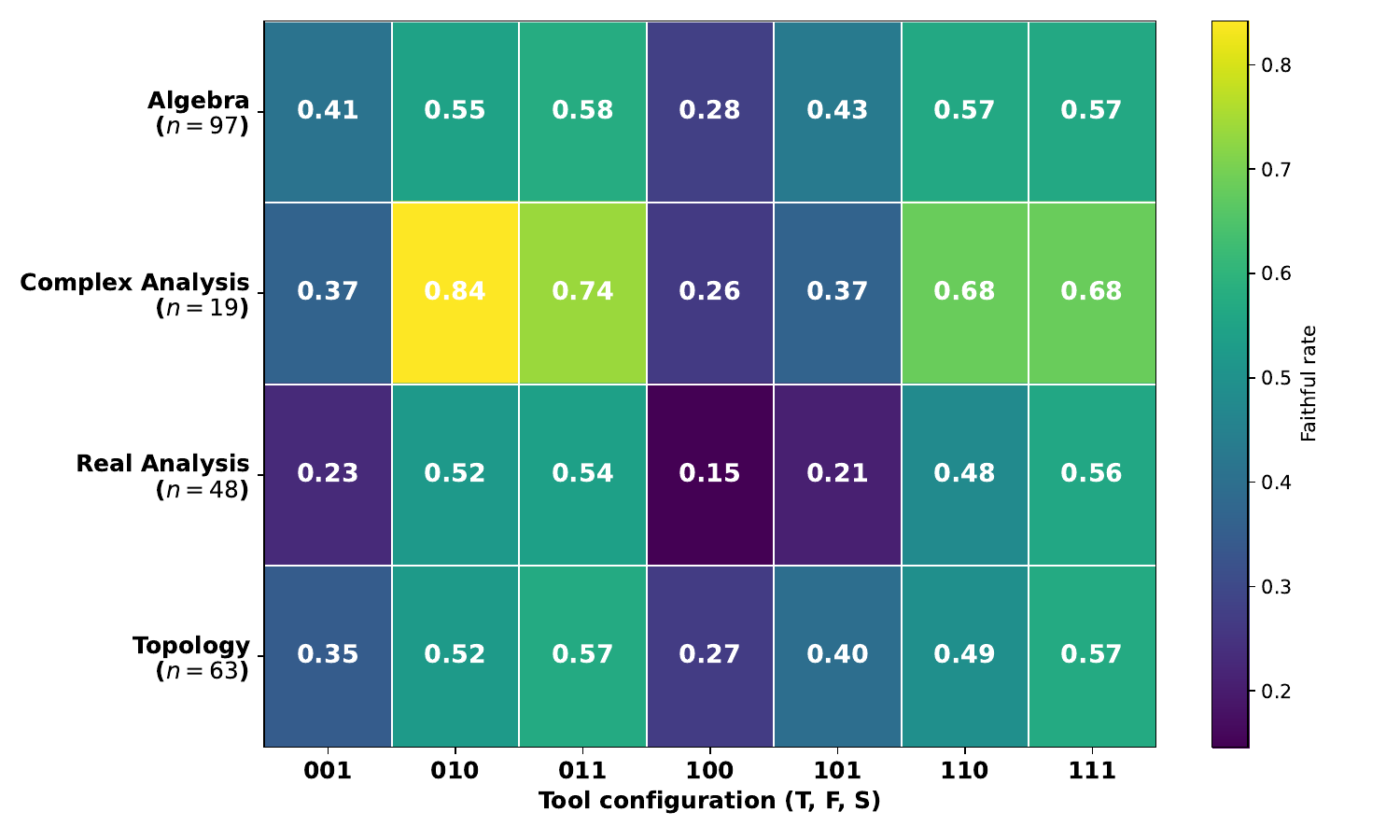}
\caption{\textbf{Faithful rate by domain and configuration.}}
\label{fig:domain_faithful_heatmap}
\end{figure}

\begin{tcolorbox}[title=Domain-wise effect of Compiler Feedback (F), colback=white, colframe=gray, breakable]
\small
\begin{table}[H]
\centering
\caption{\textbf{Domain-wise effect of compiler feedback.}
Rates average faithful outcomes over configurations with and without the feedback tool.}
\label{tab:domain_feedback_effect}
\setlength{\tabcolsep}{8pt}
\begin{tabular}{lrrr}
\toprule
\textbf{Domain} ($n$) & \textbf{F=0 (avg)} & \textbf{F=1 (avg)} & \textbf{$\Delta$F} \\
\midrule
Complex Analysis (19) & 0.26 & 0.74 & \textbf{+0.47} \\
Real Analysis (48)    & 0.17 & 0.53 & +0.36 \\
Algebra (97)          & 0.35 & 0.56 & +0.21 \\
Topology (63)         & 0.31 & 0.54 & +0.23 \\
\bottomrule
\end{tabular}
\end{table}

\noindent\textbf{Interpretation.} Complex Analysis gains most from compiler feedback (+47 pts); Algebra and Topology gain less (+21 and +23 pts).
\end{tcolorbox}

\subsection{Domain-Level Effects of Search (S)}
\label{app:domain_search}

\begin{tcolorbox}[title=Simple effect of Search conditioned on Feedback, colback=white, colframe=gray, breakable]
\small

The Search tool (S) provides symbol-level queries via \texttt{\#check}/\texttt{\#print} and general web search. It gives Mathlib information without full compilation diagnostics.

\begin{table}[H]
\centering
\caption{\textbf{Domain-wise simple effect of search conditioned on feedback.}
Search has a larger marginal effect when full Lean elaboration feedback is unavailable.}
\label{tab:domain_search_effect}
\setlength{\tabcolsep}{6pt}
\begin{tabular}{lrr|rr}
\toprule
 & \multicolumn{2}{c|}{\textbf{F=0}} & \multicolumn{2}{c}{\textbf{F=1}} \\
\textbf{Domain} & \textbf{S=0} & \textbf{S=1} & \textbf{S=0} & \textbf{S=1} \\
\midrule
Algebra (97)          & 0.28 & 0.42 & 0.56 & 0.57 \\
Complex Analysis (19) & 0.16 & 0.37 & 0.76 & 0.71 \\
Real Analysis (48)    & 0.11 & 0.22 & 0.50 & 0.55 \\
Topology (63)         & 0.24 & 0.37 & 0.51 & 0.57 \\
\midrule
\textbf{$\Delta$S, all 227} & \multicolumn{2}{c|}{\textbf{+13.9 pts}} & \multicolumn{2}{c}{\textbf{+3.1 pts}} \\
\bottomrule
\end{tabular}
\end{table}

\noindent\textbf{Interpretation.} When full REPL is unavailable (F=0), the implemented search tools provide substantial gains (+13.9 pts). Once full REPL is enabled (F=1), the effect of Search on faithfulness drops to +3.1 pts, matching the negative $F{\times}S$ interaction in Section~\ref{subsec:interaction_effects}.
\end{tcolorbox}

\subsection{Multi-Model Orchestrator Comparison}
\label{app:multi_model}

We also run our agent with Claude Sonnet 4.5 and Gemini-2.5-Pro as orchestrators on the 227 statements, with all three tools (configuration 111).

\begin{table}[H]
\centering
\caption{\textbf{Orchestrator comparison for our agent (config 111).}
Acceptance by the criterion on $N=227$.}
\label{tab:multi_model_main}
\footnotesize
\setlength{\tabcolsep}{7pt}
\begin{tabular}{lccc}
\toprule
\textbf{Orchestrator} & \textbf{One-shot} & \textbf{Agent 111} & \textbf{Uplift} \\
\midrule
GPT-5.2 & 19.8\% & 57.7\% & +37.9 pts \\
Sonnet 4.5 & 32.6\% & 63.9\% & +31.3 pts \\
Gemini-2.5-Pro & 27.3\% & 64.8\% & +37.5 pts \\
\bottomrule
\end{tabular}
\end{table}

With each orchestrator, the agent gains 71 to 86 accepted outputs over the same model's one-shot run and reaches 58--65\% acceptance, from one-shot baselines of 20--33\% (Figure~\ref{fig:multi_model} and Table~\ref{tab:multi_model_main}). The gain from tool use is thus not specific to the GPT-5.2 orchestrator.

\newpage
\end{document}